\documentclass[a4paper,11pt]{article}

\usepackage{rotating}
\usepackage{floatpag}
\usepackage{xifthen}

\rotfloatpagestyle{empty}

\ifthenelse{\isundefined{\showoptional}}{
  \newcommand{\showoptional}{1}
}{}

\ifthenelse{\isundefined{\ismain}}{
  \newcommand{\ismain}{0}
}{}

\ifthenelse{\isundefined{\lecturenotes}}{
  \newcommand{\lecturenotes}{0}
}{}

\usepackage{hyperref}

\usepackage{amsmath,amsthm,amssymb,mathabx}

\usepackage{xspace,enumerate,color,epsfig}
\usepackage{graphicx}
\graphicspath{{.}{./figures/}}

\usepackage{tikzfig}
\usepackage{stmaryrd}
\usepackage{docmute}
\usepackage{keycommand}

\newkeycommand{\wirelabel}[style=white label]{\,\tikz{\node[style=\commandkey{style}]{};}\,\xspace}

\newkeycommand{\pointmap}[style=point][1]{\,\tikz{\node[style=\commandkey{style}] (x) at (0,-0.3) {$#1$};\node [style=none] (3) at (0, 0.9) {};\node [style=none] (3) at (0, -0.9) {};\draw (x)--(0,0.7);}\,}

\newkeycommand{\point}[style=point,estyle=][1]{\,\tikz{\node[style=\commandkey{style}] (x) at (0,-0.05) {$#1$};\draw [\commandkey{estyle}] (x)--(0,1.05);}\,}

\newkeycommand{\copointmap}[style=copoint][1]{\,\tikz{\node[style=\commandkey{style}] (x) at (0,0.3) {$#1$};\draw (x)--(0,-0.7);}\,}

\newkeycommand{\dpoint}[style=point][1]{\,\tikz{\node[style=\commandkey{style},doubled] (x) at (0,-0.3) {$#1$};\draw[boldedge] (x)--(0,0.7);}\,}

\newkeycommand{\kpoint}[style=kpoint,estyle=][1]{\,\tikz{\node[style=\commandkey{style}] (x) at (0,-0.05) {$#1$};\draw [\commandkey{estyle}] (x)--(0,1.05);}\,}

\newkeycommand{\kpointgray}[style=gray kpoint,estyle=][1]{\,\tikz{\node[style=\commandkey{style}] (x) at (0,-0.05) {$#1$};\draw [\commandkey{estyle}] (x)--(0,1.05);}\,}

\newkeycommand{\typedkpoint}[style=kpoint,edgestyle=][2]{%
\,\begin{tikzpicture}
  \begin{pgfonlayer}{nodelayer}
    \node [style=none] (0) at (0, 1) {};
    \node [style=\commandkey{style}] (1) at (0, -0.75) {$#2$};
    \node [style=right label] (2) at (0.25, 0.5) {$#1$};
  \end{pgfonlayer}
  \begin{pgfonlayer}{edgelayer}
    \draw [\commandkey{edgestyle}] (1) to (0.center);
  \end{pgfonlayer}
\end{tikzpicture}\,}

\newkeycommand{\typedkpointdag}[style=kpoint adjoint,edgestyle=][2]{%
\,\begin{tikzpicture}
  \begin{pgfonlayer}{nodelayer}
    \node [style=none] (0) at (0, -1) {};
    \node [style=\commandkey{style}] (1) at (0, 0.75) {$#2$};
    \node [style=right label] (2) at (0.25, -0.5) {$#1$};
  \end{pgfonlayer}
  \begin{pgfonlayer}{edgelayer}
    \draw [\commandkey{edgestyle}] (0.center) to (1);
  \end{pgfonlayer}
\end{tikzpicture}\,}

\newkeycommand{\bistate}[style=kpoint][1]{\,\begin{tikzpicture}
  \begin{pgfonlayer}{nodelayer}
    \node [style=\commandkey{style}, minimum width=1 cm, inner sep=2pt] (0) at (0, -0.25) {$#1$};
    \node [style=none] (1) at (-0.75, 0) {};
    \node [style=none] (2) at (0.75, 0) {};
    \node [style=none] (3) at (-0.75, 0.88) {};
    \node [style=none] (4) at (0.75, 0.88) {};
  \end{pgfonlayer}
  \begin{pgfonlayer}{edgelayer}
    \draw (1.center) to (3.center);
    \draw (2.center) to (4.center);
  \end{pgfonlayer}
\end{tikzpicture}\,}

\newkeycommand{\bistatebig}[style=kpoint][1]{\,\begin{tikzpicture}
  \begin{pgfonlayer}{nodelayer}
    \node [style=\commandkey{style}, minimum width=, minimum width=1.26 cm, inner sep=2pt] (0) at (0, -0.25) {$#1$};
    \node [style=none] (1) at (-1, 0) {};
    \node [style=none] (2) at (1, 0) {};
    \node [style=none] (3) at (-1, 0.88) {};
    \node [style=none] (4) at (1, 0.88) {};
  \end{pgfonlayer}
  \begin{pgfonlayer}{edgelayer}
    \draw (1.center) to (3.center);
    \draw (2.center) to (4.center);
  \end{pgfonlayer}
\end{tikzpicture}\,}

\newkeycommand{\dbistate}[style=dkpoint][1]{\,\begin{tikzpicture}
  \begin{pgfonlayer}{nodelayer}
    \node [style=\commandkey{style}, minimum width=1 cm, inner sep=2pt] (0) at (0, -0.25) {$#1$};
    \node [style=none] (1) at (-0.75, 0) {};
    \node [style=none] (2) at (0.75, 0) {};
    \node [style=none] (3) at (-0.75, 1.0) {};
    \node [style=none] (4) at (0.75, 1.0) {};
  \end{pgfonlayer}
  \begin{pgfonlayer}{edgelayer}
    \draw [boldedge] (1.center) to (3.center);
    \draw [boldedge] (2.center) to (4.center);
  \end{pgfonlayer}
\end{tikzpicture}\,}

\newkeycommand{\bistatebraket}[style1=kpoint,style2=kpointadj][2]{\,\begin{tikzpicture}
  \begin{pgfonlayer}{nodelayer}
    \node [style=\commandkey{style1}, minimum width=1 cm, inner sep=2pt] (0) at (0, -0.75) {$#1$};
    \node [style=none] (1) at (-0.75, -0.5) {};
    \node [style=none] (2) at (0.75, -0.5) {};
    \node [style=none] (3) at (-0.75, 0.5) {};
    \node [style=none] (4) at (0.75, 0.5) {};
    \node [style=\commandkey{style2}, minimum width=1 cm, inner sep=2pt] (5) at (0, 0.75) {$#2$};
  \end{pgfonlayer}
  \begin{pgfonlayer}{edgelayer}
    \draw (1.center) to (3.center);
    \draw (2.center) to (4.center);
  \end{pgfonlayer}
\end{tikzpicture}\,}

\newkeycommand{\dkpoint}[style=dkpoint][1]{\,\tikz{\node[style=\commandkey{style}] (x) at (0,-0.1) {$#1$};\draw[boldedge] (x)--(0,1);}\,}
\newkeycommand{\dkpointadj}[style=dkpointadj][1]{\,\tikz{\node[style=\commandkey{style}] (x) at (0,0.1) {$#1$};\draw[boldedge] (x)--(0,-1);}\,}
\newkeycommand{\dkpointtrans}[style=dkpointtrans][1]{\,\tikz{\node[style=\commandkey{style}] (x) at (0,0.1) {$#1$};\draw[boldedge] (x)--(0,-1);}\,}
\newkeycommand{\dkpointconj}[style=dkpointconj][1]{\,\tikz{\node[style=\commandkey{style}] (x) at (0,-0.1) {$#1$};\draw[boldedge] (x)--(0,1);}\,}

\newcommand{\trace}{\,\begin{tikzpicture}
  \begin{pgfonlayer}{nodelayer}
    \node [style=none] (0) at (0, -0.60) {};
    \node [style=upground] (1) at (0, 0.40) {};
  \end{pgfonlayer}
  \begin{pgfonlayer}{edgelayer}
    \draw [style=boldedge] (0.center) to (1);
  \end{pgfonlayer}
\end{tikzpicture}\,}

\newcommand\discard\trace

\newkeycommand{\pointketbra}[style1=copoint,style2=point][2]{\,%
\begin{tikzpicture}
  \begin{pgfonlayer}{nodelayer}
    \node [style=none] (0) at (0, -1.75) {};
    \node [style=\commandkey{style1}] (1) at (0, -0.75) {$#1$};
    \node [style=\commandkey{style2}] (2) at (0, 0.75) {$#2$};
    \node [style=none] (3) at (0, 1.75) {};
  \end{pgfonlayer}
  \begin{pgfonlayer}{edgelayer}
    \draw (0.center) to (1);
    \draw (2) to (3.center);
  \end{pgfonlayer}
\end{tikzpicture}\,}

\newkeycommand{\twocopointketbra}[style1=copoint,style2=copoint,style3=point][3]{\,%
\begin{tikzpicture}
  \begin{pgfonlayer}{nodelayer}
    \node [style=none] (0) at (-0.75, -1.75) {};
    \node [style=none] (1) at (0.75, -1.75) {};
    \node [style=\commandkey{style1}] (2) at (-0.75, -0.75) {$#1$};
    \node [style=\commandkey{style2}] (3) at (0.75, -0.75) {$#2$};
    \node [style=\commandkey{style3}] (4) at (0, 0.75) {$#3$};
    \node [style=none] (5) at (0, 1.75) {};
  \end{pgfonlayer}
  \begin{pgfonlayer}{edgelayer}
    \draw (0.center) to (2);
    \draw (1.center) to (3);
    \draw (4) to (5.center);
  \end{pgfonlayer}
\end{tikzpicture}\,}

\newkeycommand{\twopointketbra}[style1=copoint,style2=point,style3=point][3]{\,%
\begin{tikzpicture}
  \begin{pgfonlayer}{nodelayer}
    \node [style=none] (0) at (0, -1.75) {};
    \node [style=none] (1) at (-0.75, 1.75) {};
    \node [style=\commandkey{style1}] (2) at (0, -0.75) {$#1$};
    \node [style=\commandkey{style2}] (3) at (-0.75, 0.75) {$#2$};
    \node [style=\commandkey{style3}] (4) at (0.75, 0.75) {$#3$};
    \node [style=none] (5) at (0.75, 1.75) {};
  \end{pgfonlayer}
  \begin{pgfonlayer}{edgelayer}
    \draw (0.center) to (2);
    \draw (1.center) to (3);
    \draw (4) to (5.center);
  \end{pgfonlayer}
\end{tikzpicture}\,}

\newkeycommand{\pointbraket}[style1=point,style2=copoint,estyle=][2]{\,%
\begin{tikzpicture}
  \begin{pgfonlayer}{nodelayer}
    \node [style=\commandkey{style1}] (0) at (0, -0.625) {$#1$};
    \node [style=\commandkey{style2}] (1) at (0, 0.625) {$#2$};
  \end{pgfonlayer}
  \begin{pgfonlayer}{edgelayer}
    \draw [\commandkey{estyle}] (0) to (1);
  \end{pgfonlayer}
\end{tikzpicture}\,}

\newkeycommand{\kpointketbra}[style1=kpointdag,style2=kpoint,estyle=][2]{\,%
\begin{tikzpicture}
  \begin{pgfonlayer}{nodelayer}
    \node [style=none] (0) at (0, -1.9) {};
    \node [style=\commandkey{style1}] (1) at (0, -0.95) {$#1$};
    \node [style=\commandkey{style2}] (2) at (0, 0.95) {$#2$};
    \node [style=none] (3) at (0, 1.9) {};
  \end{pgfonlayer}
  \begin{pgfonlayer}{edgelayer}
    \draw [\commandkey{estyle}] (0.center) to (1);
    \draw [\commandkey{estyle}] (2) to (3.center);
  \end{pgfonlayer}
\end{tikzpicture}\,}

\newkeycommand{\kpointmap}[style1=kpoint,style2=map][2]{\,%
\begin{tikzpicture}
  \begin{pgfonlayer}{nodelayer}
    \node [style=\commandkey{style1}] (0) at (0, -1.1) {$#1$};
    \node [style=\commandkey{style2}] (1) at (0, 0.5) {$#2$};
    \node [style=none] (2) at (0, 1.75) {};
  \end{pgfonlayer}
  \begin{pgfonlayer}{edgelayer}
    \draw (0) to (1);
    \draw (1) to (2.center);
  \end{pgfonlayer}
\end{tikzpicture}%
\,}

\newkeycommand{\dkpointmap}[style1=dkpoint,style2=dmap][2]{\,%
\begin{tikzpicture}
  \begin{pgfonlayer}{nodelayer}
    \node [style=\commandkey{style1}] (0) at (0, -1.2) {$#1$};
    \node [style=\commandkey{style2}] (1) at (0, 0.4) {$#2$};
    \node [style=none] (2) at (0, 1.7) {};
  \end{pgfonlayer}
  \begin{pgfonlayer}{edgelayer}
    \draw[style=boldedge] (0) to (1);
    \draw[style=boldedge] (1) to (2.center);
  \end{pgfonlayer}
\end{tikzpicture}%
\,}

\newkeycommand{\dkpointmapx}[style1=dkpoint,style2=dmap][2]{\,%
\begin{tikzpicture}
  \begin{pgfonlayer}{nodelayer}
    \node [style=\commandkey{style1}] (0) at (0, -1.4) {$#1$};
    \node [style=\commandkey{style2}] (1) at (0, 0.6) {$#2$};
    \node [style=none] (2) at (0, 1.9) {};
  \end{pgfonlayer}
  \begin{pgfonlayer}{edgelayer}
    \draw[style=boldedge] (0) to (1);
    \draw[style=boldedge] (1) to (2.center);
  \end{pgfonlayer}
\end{tikzpicture}%
\,}

\newkeycommand{\kpointsandwichmap}[style1=kpoint,style2=map,style3=kpointdag][3]{\,%
\begin{tikzpicture}
  \begin{pgfonlayer}{nodelayer}
    \node [style=\commandkey{style1}] (0) at (0, -1.5) {$#1$};
    \node [style=\commandkey{style2}] (1) at (0, 0) {$#2$};
    \node [style=\commandkey{style3}] (2) at (0, 1.5) {$#3$};
  \end{pgfonlayer}
  \begin{pgfonlayer}{edgelayer}
    \draw (0) to (1);
    \draw (1) to (2);
  \end{pgfonlayer}
\end{tikzpicture}%
}

\newkeycommand{\sandwichtwo}[style1=point,style2=map,style3=map,style4=copoint][4]{\,%
\begin{tikzpicture}
  \begin{pgfonlayer}{nodelayer}
    \node [style=\commandkey{style2}] (0) at (0, -1) {$#2$};
    \node [style=\commandkey{style3}] (1) at (0, 1) {$#3$};
    \node [style=\commandkey{style1}] (2) at (0, -2.6) {$#1$};
    \node [style=\commandkey{style4}] (3) at (0, 2.6) {$#4$};
  \end{pgfonlayer}
  \begin{pgfonlayer}{edgelayer}
    \draw (2) to (0);
    \draw (0) to (1);
    \draw (1) to (3);
  \end{pgfonlayer}
\end{tikzpicture}\,}

\newkeycommand{\longbraket}[style1=point,style2=copoint][2]{\,%
\begin{tikzpicture}
  \begin{pgfonlayer}{nodelayer}
    \node [style=\commandkey{style1}] (0) at (0, -2.6) {$#1$};
    \node [style=\commandkey{style2}] (1) at (0, 2.6) {$#2$};
  \end{pgfonlayer}
  \begin{pgfonlayer}{edgelayer}
    \draw (0) to (1);
  \end{pgfonlayer}
\end{tikzpicture}\,}

\newkeycommand{\onb}[style=point]{\ensuremath{\{\,\tikz{\node[style=\commandkey{style}] (x) at (0,-0.3) {$j$};\draw (x)--(0,0.7);}\,\}}\xspace}

\newkeycommand{\phase}[style=white dot][1]{\,\begin{tikzpicture}
    \begin{pgfonlayer}{nodelayer}
        \node [style=none] (0) at (0, 1) {};
        \node [style=\commandkey{style}] (2) at (0, -0) {$#1$}; 
        \node [style=none] (3) at (0, -1) {};
    \end{pgfonlayer}
    \begin{pgfonlayer}{edgelayer}
        \draw (2) to (0.center);
        \draw (3.center) to (2);
    \end{pgfonlayer}
\end{tikzpicture}\,}

\newkeycommand{\dphase}[style=white phase ddot][1]{\,\begin{tikzpicture}
    \begin{pgfonlayer}{nodelayer}
        \node [style=none] (0) at (0, 1.25) {};
        \node [style=\commandkey{style}] (2) at (0, -0) {$#1$};
        \node [style=none] (3) at (0, -1.25) {};
    \end{pgfonlayer}
    \begin{pgfonlayer}{edgelayer}
        \draw [boldedge] (2) to (0.center);
        \draw [boldedge] (3.center) to (2);
    \end{pgfonlayer}
\end{tikzpicture}\,}

\newkeycommand{\dphasepoint}[style=white phase ddot][1]{\,\begin{tikzpicture}[yshift=-3mm]
  \begin{pgfonlayer}{nodelayer}
    \node [style=none] (0) at (0, 1) {};
    \node [style=\commandkey{style}] (1) at (0, 0) {$#1$};
  \end{pgfonlayer}
  \begin{pgfonlayer}{edgelayer}
    \draw [style=boldedge] (0.center) to (1);
  \end{pgfonlayer}
\end{tikzpicture}\,}

\newkeycommand{\dphasepointgray}[style=gray phase ddot][1]{\,\begin{tikzpicture}[yshift=-3mm]
  \begin{pgfonlayer}{nodelayer}
    \node [style=none] (0) at (0, 1) {};
    \node [style=\commandkey{style}] (1) at (0, 0) {$#1$};
  \end{pgfonlayer}
  \begin{pgfonlayer}{edgelayer}
    \draw [style=boldedge] (0.center) to (1);
  \end{pgfonlayer}
\end{tikzpicture}\,}

\newkeycommand{\dphasecopoint}[style=white phase ddot][1]{\,\begin{tikzpicture}[yshift=3mm,yscale=-1]
  \begin{pgfonlayer}{nodelayer}
    \node [style=none] (0) at (0, 1) {};
    \node [style=\commandkey{style}] (1) at (0, 0) {$#1$};
  \end{pgfonlayer}
  \begin{pgfonlayer}{edgelayer}
    \draw [style=boldedge] (0.center) to (1);
  \end{pgfonlayer}
\end{tikzpicture}\,}

\newkeycommand{\dphasepointsm}[style=white phase ddot][1]{\,\begin{tikzpicture}[yshift=-3mm]
  \begin{pgfonlayer}{nodelayer}
    \node [style=none] (0) at (0, 1) {};
    \node [style=\commandkey{style}] (1) at (0, 0) {\footnotesize\!$#1$\!};
  \end{pgfonlayer}
  \begin{pgfonlayer}{edgelayer}
    \draw [style=boldedge] (0.center) to (1);
  \end{pgfonlayer}
\end{tikzpicture}\,}

\newkeycommand{\phasepoint}[style=white phase dot][1]{\,\begin{tikzpicture}[yshift=-3mm]
  \begin{pgfonlayer}{nodelayer}
    \node [style=none] (0) at (0, 1) {};
    \node [style=\commandkey{style}] (1) at (0, 0) {$#1$};
  \end{pgfonlayer}
  \begin{pgfonlayer}{edgelayer}
    \draw (0.center) to (1);
  \end{pgfonlayer}
\end{tikzpicture}\,}

\newkeycommand{\kpointbraket}[style1=kpoint,style2=kpoint adjoint,estyle=][2]{\,%
\begin{tikzpicture}
  \begin{pgfonlayer}{nodelayer}
    \node [style=\commandkey{style1}] (0) at (0, -0.735) {$#1$};
    \node [style=\commandkey{style2}] (1) at (0, 0.735) {$#2$};
  \end{pgfonlayer}
  \begin{pgfonlayer}{edgelayer}
    \draw [\commandkey{estyle}] (0) to (1);
  \end{pgfonlayer}
\end{tikzpicture}\,}

\newkeycommand{\kpointbraketx}[style1=kpoint,style2=kpoint adjoint,estyle=][2]{\,%
\begin{tikzpicture}
  \begin{pgfonlayer}{nodelayer}
    \node [style=\commandkey{style1}] (0) at (0, -0.885) {$#1$};
    \node [style=\commandkey{style2}] (1) at (0, 0.885) {$#2$};
  \end{pgfonlayer}
  \begin{pgfonlayer}{edgelayer}
    \draw [\commandkey{estyle}] (0) to (1);
  \end{pgfonlayer}
\end{tikzpicture}\,}

\tikzstyle{cdiag}=[matrix of math nodes, row sep=3em, column sep=3em, text height=1.5ex, text depth=0.25ex,inner sep=0.5em]
\tikzstyle{arrow above}=[transform canvas={yshift=0.5ex}]
\tikzstyle{arrow below}=[transform canvas={yshift=-0.5ex}]

\tikzstyle{env}=[copoint,regular polygon rotate=0,minimum width=0.2cm, fill=black]

\tikzstyle{probs}=[shape=semicircle,fill=white,draw=black,shape border rotate=180,minimum width=1.2cm]

\tikzstyle{every picture}=[baseline=-0.25em,scale=0.5]
\tikzstyle{dotpic}=[] 
\tikzstyle{diredges}=[every to/.style={diredge}]
\tikzstyle{math matrix}=[matrix of math nodes,left delimiter=(,right delimiter=),inner sep=2pt,column sep=1em,row sep=0.5em,nodes={inner sep=0pt},text height=1.5ex, text depth=0.25ex]

\tikzstyle{inline text}=[text height=1.5ex, text depth=0.25ex,yshift=0.5mm]
\tikzstyle{label}=[font=\footnotesize,text height=1.5ex, text depth=0.25ex,yshift=0.5mm]
\tikzstyle{left label}=[label,anchor=east,xshift=1.5mm]
\tikzstyle{right label}=[label,anchor=west,xshift=-1.5mm]

\tikzstyle{braceedge}=[decorate,decoration={brace,amplitude=2mm,raise=-1mm}]
\tikzstyle{small braceedge}=[decorate,decoration={brace,amplitude=1mm,raise=-1mm}]

\tikzstyle{doubled}=[line width=1.6pt] 
\tikzstyle{boldedge}=[doubled,shorten <=-0.17mm,shorten >=-0.17mm]
\tikzstyle{boldedgegray}=[doubled,gray,shorten <=-0.17mm,shorten >=-0.17mm]

\tikzstyle{semidoubled}=[line width=1.4pt] 
\tikzstyle{semiboldedgegray}=[semidoubled,gray,shorten <=-0.17mm,shorten >=-0.17mm]

\tikzstyle{boldedgedashed}=[very thick,dashed,shorten <=-0.17mm,shorten >=-0.17mm]
\tikzstyle{vboldedgedashed}=[doubled,dashed,shorten <=-0.17mm,shorten >=-0.17mm]
\tikzstyle{left hook arrow}=[left hook-latex]
\tikzstyle{right hook arrow}=[right hook-latex]
\tikzstyle{sembracket}=[line width=0.5pt,shorten <=-0.07mm,shorten >=-0.07mm]

\tikzstyle{causal edge}=[->,thick,gray]
\tikzstyle{causal nondir}=[thick,gray]
\tikzstyle{timeline}=[thick,gray, dashed]

\tikzstyle{cedge}=[<->,thick,gray!70!white]

\tikzstyle{empty diagram}=[draw=gray!40!white,dashed,shape=rectangle,minimum width=1cm,minimum height=1cm]
\tikzstyle{empty diagram small}=[draw=gray!50!white,dashed,shape=rectangle,minimum width=0.6cm,minimum height=0.5cm]

\tikzstyle{dot}=[inner sep=0mm,minimum width=2mm,minimum height=2mm,draw,shape=circle,text depth=-0.1mm]
\tikzstyle{ddot}=[inner sep=0mm, doubled, minimum width=2.5mm,minimum height=2.5mm,draw,shape=circle]

\tikzstyle{black dot}=[dot,fill=black]
\tikzstyle{white dot}=[dot,fill=white,,text depth=-0.2mm]
\tikzstyle{green dot}=[white dot] 
\tikzstyle{gray dot}=[dot,fill=gray!40!white,,text depth=-0.2mm]
\tikzstyle{red dot}=[gray dot] 

\tikzstyle{black ddot}=[ddot,fill=black]
\tikzstyle{white ddot}=[ddot,fill=white]
\tikzstyle{gray ddot}=[ddot,fill=gray!40!white]

\tikzstyle{gray edge}=[gray!40!white]

\tikzstyle{small dot}=[inner sep=0.4mm,minimum width=0pt,minimum height=0pt,draw,shape=circle]

\tikzstyle{small black dot}=[small dot,fill=black]
\tikzstyle{small white dot}=[small dot,fill=white]
\tikzstyle{small gray dot}=[small dot,fill=gray!40!white]

\tikzstyle{causal dot}=[inner sep=0.4mm,minimum width=0pt,minimum height=0pt,draw=white,shape=circle,fill=gray!40!white]

\tikzstyle{phase dimensions}=[font=\footnotesize,inner sep=0.5pt,minimum width=5mm]

\tikzstyle{white phase dot}=[dot,fill=white,phase dimensions]
\tikzstyle{white phase ddot}=[ddot,fill=white,phase dimensions]

\tikzstyle{gray phase dot}=[dot,fill=gray!40!white,phase dimensions]
\tikzstyle{gray phase ddot}=[ddot,fill=gray!40!white,phase dimensions]
\tikzstyle{grey phase dot}=[gray phase dot]
\tikzstyle{grey phase ddot}=[gray phase ddot]

\tikzstyle{cnot}=[fill=white,shape=circle,inner sep=-1.4pt]
\tikzstyle{hadamard}=[square box,inner sep=0 pt,font=\tiny\sf,minimum height=3mm,minimum width=3mm]
\tikzstyle{dhadamard}=[hadamard,doubled]
\tikzstyle{antipode}=[white dot,inner sep=0.3mm,font=\footnotesize]

\tikzstyle{scalar}=[diamond,draw,inner sep=0.5pt,font=\small]
\tikzstyle{dscalar}=[diamond,doubled, draw,inner sep=0.5pt,font=\small]

\tikzstyle{small box}=[rectangle,inline text,fill=white,draw,minimum height=5mm,yshift=-0.5mm,minimum width=5mm,font=\small]
\tikzstyle{small gray box}=[small box,fill=gray!30]
\tikzstyle{medium box}=[rectangle,inline text,fill=white,draw,minimum height=5mm,yshift=-0.5mm,minimum width=10mm,font=\small]
\tikzstyle{square box}=[small box] 
\tikzstyle{medium gray box}=[small box,fill=gray!30]
\tikzstyle{semilarge box}=[rectangle,inline text,fill=white,draw,minimum height=5mm,yshift=-0.5mm,minimum width=12.5mm,font=\small]
\tikzstyle{large box}=[rectangle,inline text,fill=white,draw,minimum height=5mm,yshift=-0.5mm,minimum width=15mm,font=\small]
\tikzstyle{large gray box}=[small box,fill=gray!30]

\tikzstyle{gray square point}=[small box,fill=gray!50]

\tikzstyle{dphase box white}=[dbox]
\tikzstyle{dphase box gray}=[dbox,fill=gray!50!white]

\tikzstyle{point}=[regular polygon,regular polygon sides=3,draw,scale=0.75,inner sep=-0.5pt,minimum width=9mm,fill=white,regular polygon rotate=180]
\tikzstyle{copoint}=[regular polygon,regular polygon sides=3,draw,scale=0.75,inner sep=-0.5pt,minimum width=9mm,fill=white]
\tikzstyle{dpoint}=[point,doubled]
\tikzstyle{dcopoint}=[copoint,doubled]

\tikzstyle{wide copoint}=[fill=white,draw,shape=isosceles triangle,shape border rotate=90,isosceles triangle stretches=true,inner sep=0pt,minimum width=1.5cm,minimum height=6.12mm]
\tikzstyle{wide point}=[fill=white,draw,shape=isosceles triangle,shape border rotate=-90,isosceles triangle stretches=true,inner sep=0pt,minimum width=1.5cm,minimum height=6.12mm,yshift=-0.0mm]
\tikzstyle{wide point plus}=[fill=white,draw,shape=isosceles triangle,shape border rotate=-90,isosceles triangle stretches=true,inner sep=0pt,minimum width=1.74cm,minimum height=7mm,yshift=-0.0mm]

\tikzstyle{wide dpoint}=[fill=white,doubled,draw,shape=isosceles triangle,shape border rotate=-90,isosceles triangle stretches=true,inner sep=0pt,minimum width=1.5cm,minimum height=6.12mm,yshift=-0.0mm]

\tikzstyle{tinypoint}=[regular polygon,regular polygon sides=3,draw,scale=0.55,inner sep=-0.15pt,minimum width=6mm,fill=white,regular polygon rotate=180] 

\tikzstyle{white point}=[point]
\tikzstyle{white dpoint}=[dpoint]
\tikzstyle{green point}=[white point] 
\tikzstyle{white copoint}=[copoint]
\tikzstyle{gray point}=[point,fill=gray!40!white]
\tikzstyle{gray dpoint}=[gray point,doubled]
\tikzstyle{red point}=[gray point] 
\tikzstyle{gray copoint}=[copoint,fill=gray!40!white]
\tikzstyle{gray dcopoint}=[gray copoint,doubled]

\tikzstyle{black point}=[point,fill=black]
\tikzstyle{black copoint}=[copoint,fill=black]

\tikzstyle{tiny gray point}=[tinypoint,fill=gray!40!white]

\tikzstyle{diredge}=[->]
\tikzstyle{rdiredge}=[<-]
\tikzstyle{thickdiredge}=[->, very thick]
\tikzstyle{pointer edge}=[->,very thick,gray]
\tikzstyle{pointer edge part}=[very thick,gray]
\tikzstyle{dashed edge}=[dashed]
\tikzstyle{thick dashed edge}=[very thick,dashed]
\tikzstyle{thick gray dashed edge}=[thick dashed edge,gray!40]
\tikzstyle{thick map edge}=[very thick,|->]

\makeatletter
\newcommand{\boxshape}[3]{%
\pgfdeclareshape{#1}{
\inheritsavedanchors[from=rectangle] 
\inheritanchorborder[from=rectangle]
\inheritanchor[from=rectangle]{center}
\inheritanchor[from=rectangle]{north}
\inheritanchor[from=rectangle]{south}
\inheritanchor[from=rectangle]{west}
\inheritanchor[from=rectangle]{east}
\backgroundpath{
\southwest \pgf@xa=\pgf@x \pgf@ya=\pgf@y
\northeast \pgf@xb=\pgf@x \pgf@yb=\pgf@y

\@tempdima=#2
\@tempdimb=#3

\pgfpathmoveto{\pgfpoint{\pgf@xa - 5pt + \@tempdima}{\pgf@ya}}
\pgfpathlineto{\pgfpoint{\pgf@xa - 5pt - \@tempdima}{\pgf@yb}}
\pgfpathlineto{\pgfpoint{\pgf@xb + 5pt + \@tempdimb}{\pgf@yb}}
\pgfpathlineto{\pgfpoint{\pgf@xb + 5pt - \@tempdimb}{\pgf@ya}}
\pgfpathlineto{\pgfpoint{\pgf@xa - 5pt + \@tempdima}{\pgf@ya}}
\pgfpathclose
}
}}

\boxshape{NEbox}{0pt}{5pt}
\boxshape{SEbox}{0pt}{-5pt}
\boxshape{NWbox}{5pt}{0pt}
\boxshape{SWbox}{-5pt}{0pt}
\boxshape{EBox}{-3pt}{3pt}
\boxshape{WBox}{3pt}{-3pt}
\makeatother

\tikzstyle{cloud}=[shape=cloud,draw,minimum width=1.5cm,minimum height=1.5cm]

\tikzstyle{map}=[draw,shape=NEbox,inner sep=2pt,minimum height=6mm,fill=white]
\tikzstyle{dashedmap}=[draw,dashed,shape=NEbox,inner sep=2pt,minimum height=6mm,fill=white]
\tikzstyle{mapdag}=[draw,shape=SEbox,inner sep=2pt,minimum height=6mm,fill=white]
\tikzstyle{mapadj}=[draw,shape=SEbox,inner sep=2pt,minimum height=6mm,fill=white]
\tikzstyle{maptrans}=[draw,shape=SWbox,inner sep=2pt,minimum height=6mm,fill=white]
\tikzstyle{mapconj}=[draw,shape=NWbox,inner sep=2pt,minimum height=6mm,fill=white]

\tikzstyle{medium map}=[draw,shape=NEbox,inner sep=2pt,minimum height=6mm,fill=white,minimum width=7mm]
\tikzstyle{medium map dag}=[draw,shape=SEbox,inner sep=2pt,minimum height=6mm,fill=white,minimum width=7mm]
\tikzstyle{medium map adj}=[draw,shape=SEbox,inner sep=2pt,minimum height=6mm,fill=white,minimum width=7mm]
\tikzstyle{medium map trans}=[draw,shape=SWbox,inner sep=2pt,minimum height=6mm,fill=white,minimum width=7mm]
\tikzstyle{medium map conj}=[draw,shape=NWbox,inner sep=2pt,minimum height=6mm,fill=white,minimum width=7mm]
\tikzstyle{semilarge map}=[draw,shape=NEbox,inner sep=2pt,minimum height=6mm,fill=white,minimum width=9.5mm]
\tikzstyle{semilarge map trans}=[draw,shape=SWbox,inner sep=2pt,minimum height=6mm,fill=white,minimum width=9.5mm]
\tikzstyle{semilarge map adj}=[draw,shape=SEbox,inner sep=2pt,minimum height=6mm,fill=white,minimum width=9.5mm]
\tikzstyle{semilarge map dag}=[draw,shape=SEbox,inner sep=2pt,minimum height=6mm,fill=white,minimum width=9.5mm]
\tikzstyle{semilarge map conj}=[draw,shape=NWbox,inner sep=2pt,minimum height=6mm,fill=white,minimum width=9.5mm]
\tikzstyle{large map}=[draw,shape=NEbox,inner sep=2pt,minimum height=6mm,fill=white,minimum width=12mm]
\tikzstyle{very large map}=[draw,shape=NEbox,inner sep=2pt,minimum height=6mm,fill=white,minimum width=17mm]

\tikzstyle{medium dmap}=[draw,doubled,shape=NEbox,inner sep=2pt,minimum height=6mm,fill=white,minimum width=7mm]
\tikzstyle{medium dmap dag}=[draw,doubled,shape=SEbox,inner sep=2pt,minimum height=6mm,fill=white,minimum width=7mm]
\tikzstyle{medium dmap adj}=[draw,doubled,shape=SEbox,inner sep=2pt,minimum height=6mm,fill=white,minimum width=7mm]
\tikzstyle{medium dmap trans}=[draw,doubled,shape=SWbox,inner sep=2pt,minimum height=6mm,fill=white,minimum width=7mm]
\tikzstyle{medium dmap conj}=[draw,doubled,shape=NWbox,inner sep=2pt,minimum height=6mm,fill=white,minimum width=7mm]
\tikzstyle{semilarge dmap}=[draw,doubled,shape=NEbox,inner sep=2pt,minimum height=6mm,fill=white,minimum width=9.5mm]
\tikzstyle{semilarge dmap trans}=[draw,doubled,shape=SWbox,inner sep=2pt,minimum height=6mm,fill=white,minimum width=9.5mm]
\tikzstyle{semilarge dmap adj}=[draw,doubled,shape=SEbox,inner sep=2pt,minimum height=6mm,fill=white,minimum width=9.5mm]
\tikzstyle{semilarge dmap dag}=[draw,doubled,shape=SEbox,inner sep=2pt,minimum height=6mm,fill=white,minimum width=9.5mm]
\tikzstyle{semilarge dmap conj}=[draw,doubled,shape=NWbox,inner sep=2pt,minimum height=6mm,fill=white,minimum width=9.5mm]
\tikzstyle{large dmap}=[draw,doubled,shape=NEbox,inner sep=2pt,minimum height=6mm,fill=white,minimum width=12mm]
\tikzstyle{large dmap conj}=[draw,doubled,shape=NWbox,inner sep=2pt,minimum height=6mm,fill=white,minimum width=12mm]
\tikzstyle{large dmap trans}=[draw,doubled,shape=SWbox,inner sep=2pt,minimum height=6mm,fill=white,minimum width=12mm]
\tikzstyle{very large dmap}=[draw,doubled,shape=NEbox,inner sep=2pt,minimum height=6mm,fill=white,minimum width=19.5mm]

\tikzstyle{muxbox}=[draw,shape=rectangle,minimum height=3mm,minimum width=3mm,fill=white]
\tikzstyle{dmuxbox}=[muxbox,doubled]

\tikzstyle{dbox}=[draw,doubled,shape=rectangle,inner sep=2pt,minimum height=6mm,minimum width=6mm,fill=white]
\tikzstyle{dmap}=[draw,doubled,shape=NEbox,inner sep=2pt,minimum height=6mm,fill=white]
\tikzstyle{dmapdag}=[draw,doubled,shape=SEbox,inner sep=2pt,minimum height=6mm,fill=white]
\tikzstyle{dmapadj}=[draw,doubled,shape=SEbox,inner sep=2pt,minimum height=6mm,fill=white]
\tikzstyle{dmaptrans}=[draw,doubled,shape=SWbox,inner sep=2pt,minimum height=6mm,fill=white]
\tikzstyle{dmapconj}=[draw,doubled,shape=NWbox,inner sep=2pt,minimum height=6mm,fill=white]

\tikzstyle{ddmap}=[draw,doubled,dashed,shape=NEbox,inner sep=2pt,minimum height=6mm,fill=white]
\tikzstyle{ddmapdag}=[draw,doubled,dashed,shape=SEbox,inner sep=2pt,minimum height=6mm,fill=white]
\tikzstyle{ddmapadj}=[draw,doubled,dashed,shape=SEbox,inner sep=2pt,minimum height=6mm,fill=white]
\tikzstyle{ddmaptrans}=[draw,doubled,dashed,shape=SWbox,inner sep=2pt,minimum height=6mm,fill=white]
\tikzstyle{ddmapconj}=[draw,doubled,dashed,shape=NWbox,inner sep=2pt,minimum height=6mm,fill=white]

\boxshape{sNEbox}{0pt}{3pt}
\boxshape{sSEbox}{0pt}{-3pt}
\boxshape{sNWbox}{3pt}{0pt}
\boxshape{sSWbox}{-3pt}{0pt}
\tikzstyle{smap}=[draw,shape=sNEbox,fill=white]
\tikzstyle{smapdag}=[draw,shape=sSEbox,fill=white]
\tikzstyle{smapadj}=[draw,shape=sSEbox,fill=white]
\tikzstyle{smaptrans}=[draw,shape=sSWbox,fill=white]
\tikzstyle{smapconj}=[draw,shape=sNWbox,fill=white]

\tikzstyle{dsmap}=[draw,dashed,shape=sNEbox,fill=white]
\tikzstyle{dsmapdag}=[draw,dashed,shape=sSEbox,fill=white]
\tikzstyle{dsmaptrans}=[draw,dashed,shape=sSWbox,fill=white]
\tikzstyle{dsmapconj}=[draw,dashed,shape=sNWbox,fill=white]

\boxshape{mNEbox}{0pt}{10pt}
\boxshape{mSEbox}{0pt}{-10pt}
\boxshape{mNWbox}{10pt}{0pt}
\boxshape{mSWbox}{-10pt}{0pt}
\tikzstyle{mmap}=[draw,shape=mNEbox]
\tikzstyle{mmapdag}=[draw,shape=mSEbox]
\tikzstyle{mmaptrans}=[draw,shape=mSWbox]
\tikzstyle{mmapconj}=[draw,shape=mNWbox]

\tikzstyle{mmapgray}=[draw,fill=gray!40!white,shape=mNEbox]
\tikzstyle{smapgray}=[draw,fill=gray!40!white,shape=sNEbox]

\makeatletter
\pgfdeclareshape{cornerpoint}{
\inheritsavedanchors[from=rectangle] 
\inheritanchorborder[from=rectangle]
\inheritanchor[from=rectangle]{center}
\inheritanchor[from=rectangle]{north}
\inheritanchor[from=rectangle]{south}
\inheritanchor[from=rectangle]{west}
\inheritanchor[from=rectangle]{east}
\backgroundpath{
\southwest \pgf@xa=\pgf@x \pgf@ya=\pgf@y
\northeast \pgf@xb=\pgf@x \pgf@yb=\pgf@y

\pgfmathsetmacro{\pgf@shorten@left}{\pgfkeysvalueof{/tikz/shorten left}}
\pgfmathsetmacro{\pgf@shorten@right}{\pgfkeysvalueof{/tikz/shorten right}}

\pgfpathmoveto{\pgfpoint{0.5 * (\pgf@xa + \pgf@xb)}{\pgf@ya - 5pt}}
\pgfpathlineto{\pgfpoint{\pgf@xa - 8pt + \pgf@shorten@left}{\pgf@yb - 1.5 * \pgf@shorten@left}}
\pgfpathlineto{\pgfpoint{\pgf@xa - 8pt + \pgf@shorten@left}{\pgf@yb}}
\pgfpathlineto{\pgfpoint{\pgf@xb + 8pt - \pgf@shorten@right}{\pgf@yb}}
\pgfpathlineto{\pgfpoint{\pgf@xb + 8pt - \pgf@shorten@right}{\pgf@yb - 1.5 * \pgf@shorten@right}}
\pgfpathclose
}
}

\pgfdeclareshape{cornercopoint}{
\inheritsavedanchors[from=rectangle] 
\inheritanchorborder[from=rectangle]
\inheritanchor[from=rectangle]{center}
\inheritanchor[from=rectangle]{north}
\inheritanchor[from=rectangle]{south}
\inheritanchor[from=rectangle]{west}
\inheritanchor[from=rectangle]{east}
\backgroundpath{
\southwest \pgf@xa=\pgf@x \pgf@ya=\pgf@y
\northeast \pgf@xb=\pgf@x \pgf@yb=\pgf@y

\pgfmathsetmacro{\pgf@shorten@left}{\pgfkeysvalueof{/tikz/shorten left}}
\pgfmathsetmacro{\pgf@shorten@right}{\pgfkeysvalueof{/tikz/shorten right}}

\pgfpathmoveto{\pgfpoint{0.5 * (\pgf@xa + \pgf@xb)}{\pgf@yb + 5pt}}
\pgfpathlineto{\pgfpoint{\pgf@xa - 8pt + \pgf@shorten@left}{\pgf@ya + 1.5 * \pgf@shorten@left}}
\pgfpathlineto{\pgfpoint{\pgf@xa - 8pt + \pgf@shorten@left}{\pgf@ya}}
\pgfpathlineto{\pgfpoint{\pgf@xb + 8pt - \pgf@shorten@right}{\pgf@ya}}
\pgfpathlineto{\pgfpoint{\pgf@xb + 8pt - \pgf@shorten@right}{\pgf@ya + 1.5 * \pgf@shorten@right}}
\pgfpathclose
}
}

\makeatother

\pgfkeyssetvalue{/tikz/shorten left}{0pt}
\pgfkeyssetvalue{/tikz/shorten right}{0pt}

\tikzstyle{kpoint common}=[draw,fill=white,inner sep=1pt,minimum height=4mm]
\tikzstyle{kpoint}=[shape=cornerpoint,shorten left=5pt,kpoint common]
\tikzstyle{kpoint adjoint}=[shape=cornercopoint,shorten left=5pt,kpoint common]
\tikzstyle{kpoint conjugate}=[shape=cornerpoint,shorten right=5pt,kpoint common]
\tikzstyle{kpoint transpose}=[shape=cornercopoint,shorten right=5pt,kpoint common]
\tikzstyle{kpoint symm}=[shape=cornerpoint,shorten left=5pt,shorten right=5pt,kpoint common]

\tikzstyle{kpointdag}=[kpoint adjoint]
\tikzstyle{kpointadj}=[kpoint adjoint]
\tikzstyle{kpointconj}=[kpoint conjugate]
\tikzstyle{kpointtrans}=[kpoint transpose]

\tikzstyle{big kpoint}=[kpoint, minimum width=1.2 cm, minimum height=8mm, inner sep=4pt, text depth=3mm]

\tikzstyle{wide kpoint}=[kpoint, minimum width=1 cm, inner sep=2pt, text depth=-0.7 mm]
\tikzstyle{wide kpointdag}=[kpointdag, minimum width=1 cm, inner sep=2pt, text depth=0.7 mm]
\tikzstyle{wide kpointconj}=[kpointconj, minimum width=1 cm, inner sep=2pt, text depth=-0.7 mm]
\tikzstyle{wide kpointtrans}=[kpointtrans, minimum width=1 cm, inner sep=2pt, text depth=0.7 mm]

\tikzstyle{gray kpoint}=[kpoint,fill=gray!50!white]
\tikzstyle{gray kpointdag}=[kpointdag,fill=gray!50!white]
\tikzstyle{gray kpointadj}=[kpointadj,fill=gray!50!white]
\tikzstyle{gray kpointconj}=[kpointconj,fill=gray!50!white]
\tikzstyle{gray kpointtrans}=[kpointtrans,fill=gray!50!white]

\tikzstyle{gray dkpoint}=[kpoint,fill=gray!50!white,doubled]
\tikzstyle{gray dkpointdag}=[kpointdag,fill=gray!50!white,doubled]
\tikzstyle{gray dkpointadj}=[kpointadj,fill=gray!50!white,doubled]
\tikzstyle{gray dkpointconj}=[kpointconj,fill=gray!50!white,doubled]
\tikzstyle{gray dkpointtrans}=[kpointtrans,fill=gray!50!white,doubled]

\tikzstyle{white label}=[draw,fill=white,rectangle,inner sep=0.7 mm]
\tikzstyle{gray label}=[draw,fill=gray!50!white,rectangle,inner sep=0.7 mm]
\tikzstyle{black label}=[draw,fill=black,rectangle,inner sep=0.7 mm]

\tikzstyle{dkpoint}=[kpoint,doubled]
\tikzstyle{wide dkpoint}=[wide kpoint,doubled]
\tikzstyle{dkpointdag}=[kpoint adjoint,doubled]
\tikzstyle{dkcopoint}=[kpoint adjoint,doubled]
\tikzstyle{dkpointadj}=[kpoint adjoint,doubled]
\tikzstyle{dkpointconj}=[kpoint conjugate,doubled]
\tikzstyle{dkpointtrans}=[kpoint transpose,doubled]

\tikzstyle{kscalar}=[kpoint common, shape=EBox, inner xsep=-1pt, inner ysep=3pt,font=\small]
\tikzstyle{kscalarconj}=[kpoint common, shape=WBox, inner xsep=-1pt, inner ysep=3pt,font=\small]

 \tikzstyle{upground}=[circuit ee IEC,thick,ground,rotate=90,scale=2.5]
 \tikzstyle{downground}=[circuit ee IEC,thick,ground,rotate=-90,scale=2.5]
 \tikzstyle{bigground}=[regular polygon,regular polygon sides=3,draw=gray,scale=0.50,inner sep=-0.5pt,minimum width=10mm,fill=gray]

\tikzstyle{arrs}=[-latex,font=\small,auto]
\tikzstyle{arrow plain}=[arrs]
\tikzstyle{arrow dashed}=[dashed,arrs]
\tikzstyle{arrow bold}=[very thick,arrs]
\tikzstyle{arrow hide}=[draw=white!0,-]
\tikzstyle{arrow reverse}=[latex-]
\tikzstyle{cdnode}=[]

\let\olddagger\dagger
\renewcommand{\dagger}{\ensuremath{\olddagger}\xspace}

\usepackage{makeidx}
\makeindex

\theoremstyle{definition}
\newtheorem{theorem}{Theorem}[section]
\newtheorem{definition}[theorem]{Definition}

\newcommand{\TODO}[1]{\marginpar{\scriptsize\bB \textbf{TODO:} #1\e}}

\newcommand{\TODOa}[1]{\marginpar{\scriptsize\bM \textbf{TODO:} #1\e}}
\newcommand{\TODOb}[1]{\marginpar{\scriptsize\bB \textbf{TODO:} #1\e}}

\newcommand{\COMMa}[1]{\marginpar{\scriptsize\bM \textbf{COMM:} #1\e}}
\newcommand{\COMMb}[1]{\marginpar{\scriptsize\bB \textbf{COMM:} #1\e}}

\newcommand{\CHECK}[1]{\marginpar{\scriptsize\bR \textbf{CHECK:} #1\e}}

\usepackage[color]{changebar}

\usepackage{color}
\def\bR{\begin{color}{red}} 
\def\bB{\begin{color}{blue}}
\def\bM{\begin{color}{magenta}}
\def\bC{\begin{color}{cyan}}
\def\bW{\begin{color}{white}}
\def\bBl{\begin{color}{black}} 
\def\bG{\begin{color}{green}}
\def\bY{\begin{color}{yellow}}
\def\e{\end{color}\xspace}
\newcommand{\bit}{\begin{itemize}}
\newcommand{\eit}{\end{itemize}\par\noindent}
\newcommand{\ben}{\begin{enumerate}}
\newcommand{\een}{\end{enumerate}\par\noindent}
\newcommand{\beq}{\begin{equation}}
\newcommand{\eeq}{\end{equation}\par\noindent}
\newcommand{\beqa}{\begin{eqnarray*}}
\newcommand{\eeqa}{\end{eqnarray*}\par\noindent}
\newcommand{\beqn}{\begin{eqnarray}}
\newcommand{\eeqn}{\end{eqnarray}\par\noindent}

\if\lecturenotes1

\renewcommand{\TODO}[1]{}
\renewcommand{\TODOa}[1]{}
\renewcommand{\TODOb}[1]{}
\renewcommand{\COMMa}[1]{}
\renewcommand{\COMMb}[1]{}
\renewcommand{\CHECK}[1]{}

\def\bR{\begin{color}{black}} 
\def\bB{\begin{color}{black}}
\def\bM{\begin{color}{black}}
\def\bC{\begin{color}{black}}
\def\bW{\begin{color}{black}}
\def\bG{\begin{color}{black}}
\def\bY{\begin{color}{black}}

\fi

\title{From quantum foundations via natural\\ language meaning to a theory of everything}         
\author{Bob Coecke\\ University of Oxford\\  coecke@cs.ox.ac.uk}  
\date{}

\begin{document} 
\maketitle
\begin{abstract} 
In this paper we argue for a paradigmatic shift from `reductionism' to `togetherness'.  In particular, we show how interaction between systems in quantum theory naturally carries over to modelling how word meanings interact in natural language. Since meaning in natural language, depending on the subject domain, encompasses discussions within any scientific discipline, we obtain a template for theories such as social interaction, animal behaviour, and many others.     
\end{abstract}

\section{...in the beginning was $\otimes$}

No  physicists! ...the symbol $\otimes$  above does not stand  for the operation that turns two Hilbert spaces into the smallest Hilbert space in which the two given ones bilinearly embed.  No  category-theoreticians! ...neither does it stand for the composition operation that turns any pair of objects (and morphisms) in a monoidal category into another object, and that is subject to a horrendous bunch of conditions that guaranty coherence with the remainder of the structure.   Instead, this is what it means:  
\[
\otimes \equiv \mbox{``togetherness''}   
\]
More specifically, it represents the togetherness of foo${}_1$ and foo${}_2$ without giving any specification of who/what foo${}_1$ and foo${}_2$ actually are.  Differently put, it's the new stuff that  emerges when foo${}_1$ and foo${}_2$ get together.  If they don't like each other at all, this may be a fight. If they do like each other a lot, this may be a marriage, and a bit later, babies.   Note that togetherness is  vital for the emergence to actually take place, given that it is quite hard to either have a fight, a wedding, or a baby,  if there is nobody else around. 

It is of course true that in von Neumann's formalisation of  quantum theory the \em tensor product of Hilbert spaces \em (also denoted by $\otimes$) plays this role \cite{vN}, giving rise to the emergent phenomenon of \em entanglement \em \cite{EPR, Schrodinger}. And more generally, in category theory one can axiomatise \em composition of objects \em  (again denoted by $\otimes$)  within a \em symmetric monoidal category \em \cite{benabou}, giving rise to elements that don't simply arise by pairing, just like in the case of the Hilbert space tensor product.    

However, in the  case of von Neumann's formalisation of  quantum theory we are talking about a formalisation which, despite being widely used, its creator von Neumann himself didn't even like \cite{Redei1}. Moreover, in this formalism $\otimes$ only arises as a secondary construct,  requiring a detailed description of  foo${}_1$ and foo${}_2$, whose togetherness it describes.   What we are after is a `foo-less' conception of $\otimes$. The composition operation $\otimes$ in symmetric monoidal categories heads in that direction.  However,  by making an unnecessary commitment to set-theory, it makes things unnecessarily complicated \cite{CatsII}.  Moreover, while this operation is general enough to accommodate togetherness, it  doesn't really tell us anything about it. 
 
The  title of this section  
is a metaphor aimed at confronting 
 the complete disregard that the concept of togetherness has suffered in the sciences, and especially, in physics, where all of the effort has been on describing the individual, typically by breaking its description down to that of even smaller individuals.  While, without any doubt, this has been a useful endeavour, it unfortunately has evolved in a rigid doctrine, leaving no space for anything else. The most extreme manifestation of this dogma is the use of the term `theory of everything' in particle physics.   We will provide an alternative conceptual template for a theory of everything, supported not only by scientific examples, but also by everyday ones. 

Biology evolved from chopping up individual animals in laboratories, to considering them in the context of other other animals and varying environments.  the result is the theory of evolution of species. Similarly, our current (still very poor) understanding of the human brain makes it clear that the human brain should  not be studied as something in isolation, but as something that fundamentally requires interaction with other brains \cite{lieberman2013social}.  In contemporary audio equipment, music consists of nothing  but a strings of zeros and ones.  Instead, the entities that truly make up music are pitch, sound, rhythm, chord progression, crescendo, and so on.  And in particular, music is not just a bag of these, since their intricate interaction  is even more important than these constituents themselves.  The same is true for film, where it isn't even that clear  what it is made up from, but it does include such things as (easily replaceable) actors, decors, cameras, which all are  part of a soup stirred by a director.  But again,  in contemporary video equipment, it is nothing but a string of zeros and ones. 

In fact, everything that goes on in  pretty much all  modern devices is  nothing but zeros and ones. While it was Turing's brilliance to realise that this could in fact be done, and provided a foundation for the theory of computability \cite{Turing}, this is in fact the only place where the zeros and ones are truly meaningful, in the form of a Turing machine.  Elsewhere, it is nothing but a (universal) representation, with no conceptual  qualities regarding the subject matter.       

\section{Formalising togetherness 1: not  there yet}  

So, how does one go about formalising the concept of togetherness?  While we don't want  an explicit description of the foo involved,   we do need some kind of means for identifying foo.  Therefore, we  simple give each foo a name, say $A, B, C, \ldots$.  Then, $A\otimes B$ represents the togetherness of $A$ and $B$.  We also don't want an explicit description of $A\otimes B$, so how can we say anything about $\otimes$ without explicitly describing $A$,  $B$ and $A\otimes B$?  

Well, rather than describing these systems themselves, we could describe their relationships. For example, in a certain theory togetherness could obey the following equation:
\[
A\otimes A = A
\]
That is, togetherness of two copies of something is exactly the same as a single copy, or in simpler terms, one is as good as two.  For example, if one is in need of a plumber to fix a pipe, one only needs one.  The only thing a second plumber would contribute is a bill for the time he wasted coming to your house.  Obviously, this is not the kind of togetherness  that we are really interested in, given that this kind adds nothing at all.  

A tiny bit more interesting is the case that two is as good as three:
\[
A\otimes A\otimes A = A\otimes A
\]
e.g.~when something needs to be carried on a staircase, but there really is only space for two people to be involved.  Or, when $A$ is female and $\bar A$ is male, and the goal is reproduction, we have:
\[
A\otimes \bar A\otimes \bar A = A\otimes \bar A
\]
(ignoring  testosterone induced scuffles and the benefits of natural selection.)  

We really won't get very far this manner.  One way in which things can be improved is by replacing equations by inequalities.  For example, while:
\[
A = B
\]
simply means that one of the two is redundant, instead:
\[
A \leq B
\]
can mean that from $A$ we can produce $B$, and:
\[
A \otimes B \leq C
\]
can mean that from $A$ and $B$ together we can produce $C$, and:
\[
A \otimes C \leq B \otimes C
\]
can mean that in the presence of $C$ from $A$ we can produce $B$, i.e.~that $C$ is a catalyst.    

What we have now is a so-called \em resource theory\em, that is, a theory which captures how  stuff we care about can be interconverted \cite{CFS}.  Resource theories allow for quantitative analysis, for example, in terms of a \em conversion rate\em:
\[
r(A\to B) := {\rm sup}\Biggl\{ {m\over n} \Biggm| \underbrace{A\otimes \ldots\otimes A}_n \leq \underbrace{B\otimes \ldots\otimes B}_m\Biggr\}
\]
So evidently we have some  genuine substance now.\footnote{In fact, resource theories are currently a very active area of research in the quantum information and quantum foundations communities, e.g.~the resource theories of entanglement \cite{EntanglementResource}, symmetry \cite{gour2008resource}, and athermality \cite{brandao2011resource}.} 

\section{Formalising togetherness 2: that's better}       

But we can still do a lot better.  What a resource theory fails to capture (on purpose in fact) is the actual process that converts one resource into another one.  So let's fix that problem, and explicitly account for processes. 

In terms of togetherness, this means that we bring the fun foo${}_1$ and foo${}_2$ can have together  explicitly in the picture. Let:
\[
f: A\to B 
\]
denote some process that transforms  $A$ into  $B$.Then, given two such processes $f_1$ and $f_2$ we can also consider their togetherness: 
\[
f_1\otimes f_2: A_1\otimes A_2\to B_1\otimes B_2
\]
Moreover, some processes can be sequentially chained:  
\[
g\circ f: A\to B \to C
\]
We say `some', since  $f$ has to produce $B$, in order for:    
\[
g:B\to C
\]
to take place.

Now, here one may end up in a bit of a mess if one isn't clever.  In particular, with a bit of thinking one quickly realises that one wants some equations to be obeyed, for example:
\beq\label{eq:alg1a}
(f_1\otimes f_2)\otimes f_3 = f_1\otimes (f_2\otimes f_3)  
\eeq
\beq\label{eq:alg1b}
 h\circ(g\circ f) = (h\circ g)\circ f
\eeq
and a bit more sophisticated, also:
\beq\label{eq:alg2}
(g_1\otimes g_2)\circ(f_1\otimes f_2)=(g_1\circ f_1)\otimes (g_2\circ f_2)
\eeq
There may even be some more equations that one wants to have, but which ones?  This turns out to be a very difficult problem.  Too difficult in the light of our limited existence in this world. The origin of this problem is that we treat $\otimes$, and also $\circ$, as  algebraic connectives, and that algebra has its roots in set-theory.  The larger-than-life problem can be avoided in a manner that is equally elegant as it is simple.  

To state that things are together, we just write them down together:    
\[
A\qquad B
\]
There really is no reason to adjoin the symbol $\otimes$ between them.  Now, this $A$ and $B$ will play the role of an input or an output  of processes transforming them. Therefore, it will be useful to represent them by a wire:
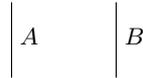
\[
\begin{tikzpicture}
	\begin{pgfonlayer}{nodelayer}
		\node [style=none] (0) at (-1.25, 1) {};
		\node [style=none] (1) at (-1.25, -1) {};
		\node [style=none] (2) at (1.5, 1) {};
		\node [style=none] (3) at (1.5, -1) {};
		\node [style=right label] (4) at (1.75, 0) {$B$};
		\node [style=right label] (5) at (-1, 0) {$A$};
	\end{pgfonlayer}
	\begin{pgfonlayer}{edgelayer}
		\draw [style=swap] (0.center) to (1.center);
		\draw [style=swap] (2.center) to (3.center);
	\end{pgfonlayer}
\end{tikzpicture} 
\]
Then, a process transforming $A$ into $B$ can be represented by a box: 
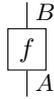
\[
\begin{tikzpicture}
	\begin{pgfonlayer}{nodelayer}
		\node [style=none] (0) at (0, 1.25) {};
		\node [style=small box] (1) at (0, 0) {$f$};
		\node [style=none] (2) at (0, -1.25) {};
		\node [style=right label] (3) at (0.25, -1) {$A$};
		\node [style=right label, yshift=-0.6mm] (4) at (0.25, 1) {$B$};
	\end{pgfonlayer}
	\begin{pgfonlayer}{edgelayer}
		\draw [style=swap] (0.center) to (1);
		\draw [style=swap] (1) to (2.center);
	\end{pgfonlayer}
\end{tikzpicture}
\]
Togetherness of processes now becomes:
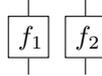
\[
\begin{tikzpicture}
	\begin{pgfonlayer}{nodelayer}
		\node [style=none] (0) at (0.75, 1) {};
		\node [style=none] (1) at (0.75, -1) {};
		\node [style=none] (2) at (-0.75, 1) {};
		\node [style=none] (3) at (-0.75, -1) {};
		\node [style=small box] (4) at (-0.75, 0) {$f_1\!$};
		\node [style=small box] (5) at (0.75, 0) {$f_2\!$};
	\end{pgfonlayer}
	\begin{pgfonlayer}{edgelayer}
		\draw [style=swap] (4) to (3.center);
		\draw [style=swap] (2.center) to (4);
		\draw [style=swap] (5) to (1.center);
		\draw [style=swap] (0.center) to (5);
	\end{pgfonlayer}
\end{tikzpicture}
\]
 and  chaining processes becomes:
\[
\begin{tikzpicture}
	\begin{pgfonlayer}{nodelayer}
		\node [style=small box] (0) at (0, -0.75) {$f$};
		\node [style=small box] (1) at (0, 0.75) {$g$};
		\node [style=none] (2) at (0, -1.75) {};
		\node [style=none] (3) at (0, 1.75) {};
	\end{pgfonlayer}
	\begin{pgfonlayer}{edgelayer}
		\draw [style=swap] (0) to (2.center);
		\draw [style=swap] (1) to (0);
		\draw [style=swap] (3.center) to (1);
	\end{pgfonlayer}
\end{tikzpicture}
\]
In particular, equations (\ref{eq:alg1a}),  (\ref{eq:alg1b}) and (\ref{eq:alg2}) become:   
\[
\begin{tikzpicture}
	\begin{pgfonlayer}{nodelayer}
		\node [style=none] (0) at (5.75, 4.5) {};
		\node [style=small box] (1) at (5.75, 0.5) {$f$};
		\node [style=none] (2) at (5.75, -0.5) {};
		\node [style=small box] (3) at (5.75, 2) {$g$};
		\node [style=small box] (4) at (5.75, 3.5) {$h$};
		\node [style=small box] (5) at (8.75, 2) {$g$};
		\node [style=small box] (6) at (8.75, 3.5) {$h$};
		\node [style=none] (7) at (8.75, 4.5) {};
		\node [style=none] (8) at (8.75, -0.5) {};
		\node [style=small box] (9) at (8.75, 0.5) {$f$};
		\node [style=none] (10) at (7.25, 2) {$=$};
		\node [style=small box] (11) at (-8.75, 2) {$f_1\!$};
		\node [style=none] (12) at (-8.75, 1) {};
		\node [style=none] (13) at (-8.75, 3) {};
		\node [style=none] (14) at (-4.25, 2) {$=$};
		\node [style=small box] (15) at (-3, -2.5) {$g_1\!$};
		\node [style=none] (16) at (-3, -5) {};
		\node [style=small box] (17) at (-3, -4) {$f_1\!$};
		\node [style=none] (18) at (-3, -1.5) {};
		\node [style=none] (19) at (-1.5, -5) {};
		\node [style=small box] (20) at (-1.5, -2.5) {$g_2\!$};
		\node [style=small box] (21) at (-1.5, -4) {$f_2\!$};
		\node [style=none] (22) at (-1.5, -1.5) {};
		\node [style=none] (23) at (3, -5) {};
		\node [style=none] (24) at (1.5, -5) {};
		\node [style=small box] (25) at (1.5, -4) {$f_1\!$};
		\node [style=small box] (26) at (1.5, -2.5) {$g_1\!$};
		\node [style=none] (27) at (3, -1.5) {};
		\node [style=small box] (28) at (3, -4) {$f_2\!$};
		\node [style=none] (29) at (1.5, -1.5) {};
		\node [style=small box] (30) at (3, -2.5) {$g_2\!$};
		\node [style=none] (31) at (0, -3.25) {$=$};
		\node [style=none] (32) at (-7.25, 1) {};
		\node [style=none] (33) at (-7.25, 3) {};
		\node [style=small box] (34) at (-7.25, 2) {$f_2\!$};
		\node [style=none] (35) at (-5.75, 1) {};
		\node [style=none] (36) at (-5.75, 3) {};
		\node [style=small box] (37) at (-5.75, 2) {$f_3\!$};
		\node [style=none] (38) at (-1.25, 3) {};
		\node [style=none] (39) at (-2.75, 3) {};
		\node [style=none] (40) at (-2.75, 1) {};
		\node [style=small box] (41) at (-2.75, 2) {$f_1\!$};
		\node [style=small box] (42) at (-1.25, 2) {$f_2\!$};
		\node [style=none] (43) at (0.25, 3) {};
		\node [style=none] (44) at (-1.25, 1) {};
		\node [style=small box] (45) at (0.25, 2) {$f_3\!$};
		\node [style=none] (46) at (0.25, 1) {};
	\end{pgfonlayer}
	\begin{pgfonlayer}{edgelayer}
		\draw [style=swap] (1) to (2.center);
		\draw [style=swap] (0.center) to (4);
		\draw [style=swap] (4) to (3);
		\draw [style=swap] (3) to (1);
		\draw [style=swap] (9) to (8.center);
		\draw [style=swap] (7.center) to (6);
		\draw [style=swap] (6) to (5);
		\draw [style=swap] (5) to (9);
		\draw [style=swap] (11) to (12.center);
		\draw [style=swap] (13.center) to (11);
		\draw [style=swap] (17) to (16.center);
		\draw [style=swap] (15) to (17);
		\draw [style=swap] (18.center) to (15);
		\draw [style=swap] (21) to (19.center);
		\draw [style=swap] (20) to (21);
		\draw [style=swap] (22.center) to (20);
		\draw [style=swap] (25) to (24.center);
		\draw [style=swap] (26) to (25);
		\draw [style=swap] (29.center) to (26);
		\draw [style=swap] (28) to (23.center);
		\draw [style=swap] (30) to (28);
		\draw [style=swap] (27.center) to (30);
		\draw [style=swap] (34) to (32.center);
		\draw [style=swap] (33.center) to (34);
		\draw [style=swap] (37) to (35.center);
		\draw [style=swap] (36.center) to (37);
		\draw [style=swap] (41) to (40.center);
		\draw [style=swap] (39.center) to (41);
		\draw [style=swap] (42) to (44.center);
		\draw [style=swap] (38.center) to (42);
		\draw [style=swap] (45) to (46.center);
		\draw [style=swap] (43.center) to (45);
	\end{pgfonlayer}
\end{tikzpicture}
\]
That is,  all equations have become tautologies!\footnote{A more extensive discussion of this bit of magic can be found in \cite{CatsII, Gospel, CKbook}.}  

\section{Anti-cartesian togetherness}  

One important kind of processes are \em states\em:
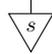
\[
\begin{tikzpicture}
	\begin{pgfonlayer}{nodelayer}
		\node [style=point] (0) at (0, -0.25) {$s$};
		\node [style=none] (1) at (0, 0.5) {};
	\end{pgfonlayer}
	\begin{pgfonlayer}{edgelayer}
		\draw [style=swap] (1.center) to (0);
	\end{pgfonlayer}
\end{tikzpicture}
\]
These are depicted without any inputs, where `no wire' can be read as `nothing' (or `no-foo').\footnote{That we use triangles for these rather than boxes is inspired by the Dirac notation which is used in quantum theory. Please consult  \cite{Kindergarten, CatsII, CKbook} for a discussion.}  
The opposite notion is that of an \em effect\em, that is, a process without an output:
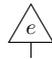
\[
\begin{tikzpicture}
	\begin{pgfonlayer}{nodelayer}
		\node [style=copoint] (0) at (0, 0.25) {$e$};
		\node [style=none] (1) at (0, -0.5) {};
	\end{pgfonlayer}
	\begin{pgfonlayer}{edgelayer}
		\draw [style=swap] (1.center) to (0);
	\end{pgfonlayer}
\end{tikzpicture}
\]
borrowing terminology  from quantum theory.\footnote{Examples of these include `tests' \cite{CKbook}.}

We can now identify those theories in which togetherness \underline{doesn't} yield anything new.  Life in such a world is pretty lonely...

\begin{definition}
A theory of togetherness is \em cartesian \em if each state:    
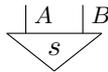
\[
\begin{tikzpicture}
	\begin{pgfonlayer}{nodelayer}
		\node [style=right label, yshift=-0.6mm] (0) at (1, 0.5) {$B$};
		\node [style=right label, yshift=-0.6mm] (1) at (-0.5, 0.5) {$A$};
		\node [style=none] (2) at (-1.25, 0) {};
		\node [style=none] (3) at (-0.75, 0) {};
		\node [style=none] (4) at (1.25, 0) {};
		\node [style=none] (5) at (0, -1) {};
		\node [style=none] (6) at (-0.75, 0.75) {};
		\node [style=none, yshift=1.2pt] (7) at (0, -0.5) {$s$}; 
		\node [style=none] (8) at (0.75, 0) {};
		\node [style=none] (9) at (0.75, 0.75) {};
	\end{pgfonlayer}
	\begin{pgfonlayer}{edgelayer}
		\draw [style=swap] (6.center) to (3.center);
		\draw [style=swap] (9.center) to (8.center);
		\draw [style=swap] (2.center) to (5.center);
		\draw [style=swap] (5.center) to (4.center);
		\draw [style=swap] (4.center) to (2.center);
	\end{pgfonlayer}
\end{tikzpicture}
\]
decomposes as follows: 
\[
\begin{tikzpicture}
	\begin{pgfonlayer}{nodelayer}
		\node [style=right label, yshift=-0.6mm] (0) at (1, 0.5) {$B$};
		\node [style=right label, yshift=-0.6mm] (1) at (-0.5, 0.5) {$A$};
		\node [style=none] (2) at (-1.25, 0) {};
		\node [style=none] (3) at (-0.75, 0) {};
		\node [style=none] (4) at (1.25, 0) {};
		\node [style=none] (5) at (0, -1) {};
		\node [style=none] (6) at (-0.75, 0.75) {};
		\node [style=none, yshift=1.2pt] (7) at (0, -0.5) {$s$}; 
		\node [style=none] (8) at (0.75, 0) {};
		\node [style=none] (9) at (0.75, 0.75) {};
	\end{pgfonlayer}
	\begin{pgfonlayer}{edgelayer}
		\draw [style=swap] (6.center) to (3.center);
		\draw [style=swap] (9.center) to (8.center);
		\draw [style=swap] (2.center) to (5.center);
		\draw [style=swap] (5.center) to (4.center);
		\draw [style=swap] (4.center) to (2.center);
	\end{pgfonlayer}
\end{tikzpicture}\ = \ \ \raisebox{1mm}{\begin{tikzpicture}
	\begin{pgfonlayer}{nodelayer}
		\node [style=point] (0) at (-0.75, -0.5) {$s_1$};
		\node [style=none] (1) at (-0.75, 0.5) {};
		\node [style=none] (2) at (0.75, 0.5) {};
		\node [style=point] (3) at (0.75, -0.5) {$s_2$};
		\node [style=right label, yshift=-0.6mm] (4) at (-0.5, 0.25) {$A$};
		\node [style=right label, yshift=-0.6mm] (5) at (1, 0.25) {$B$};
	\end{pgfonlayer}
	\begin{pgfonlayer}{edgelayer}
		\draw [style=swap] (1.center) to (0);
		\draw [style=swap] (2.center) to (3);
	\end{pgfonlayer}
\end{tikzpicture}}
\]
\end{definition}

So cartesianness means that all possible realisations of two foo-s can be achieved by pairing  realisations of the individual  foo-s involved. In short,  a whole can be described in term of its parts, rendering togetherness a void concept.  So very lonely and indeed... But, wait a minute.  Why is it then the case that so much of traditional mathematics follows this cartesian template, and that even category theory for a long time has followed a strict cartesian stance?  Beats me.  Seriously...beats me! 

Anyway, an obvious consequence of this is that for those areas where togetherness is a genuinely non-trivial concept, traditional mathematical structures  aren't always that useful.  That is maybe why social sciences don't make much  use of any kind of modern pure mathematics.   

And now for something completely different:  

\begin{definition}
A theory of togetherness is  \em anti-cartesian \em if for each $A$ there exists $A^*$, a special state $\cup$ and a special effect $\cap$:
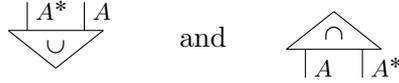
\[
\begin{tikzpicture}
	\begin{pgfonlayer}{nodelayer}
		\node [style=right label, yshift=-0.6mm] (0) at (1, 0.75) {$A$};
		\node [style=right label, yshift=-0.6mm] (1) at (-0.5, 0.75) {$A^*$};
		\node [style=none] (2) at (-1.25, 0.25) {};
		\node [style=none] (3) at (-0.75, 0.25) {};
		\node [style=none] (4) at (1.25, 0.25) {};
		\node [style=none] (5) at (0, -0.75) {};
		\node [style=none] (6) at (-0.75, 1) {};
		\node [style=none, yshift=1.2pt] (7) at (0, -0.25) {$\cup$};
		\node [style=none] (8) at (0.75, 0.25) {};
		\node [style=none] (9) at (0.75, 1) {};
	\end{pgfonlayer}
	\begin{pgfonlayer}{edgelayer}
		\draw [style=swap] (6.center) to (3.center);
		\draw [style=swap] (9.center) to (8.center);
		\draw [style=swap] (2.center) to (5.center);
		\draw [style=swap] (5.center) to (4.center);
		\draw [style=swap] (4.center) to (2.center);
	\end{pgfonlayer}
\end{tikzpicture}\qquad\mbox{and}\qquad\begin{tikzpicture}
	\begin{pgfonlayer}{nodelayer}
		\node [style=right label, yshift=-0.6mm] (0) at (1, -0.75) {$A^*$};
		\node [style=right label, yshift=-0.6mm] (1) at (-0.5, -0.75) {$A$};
		\node [style=none] (2) at (-1.25, -0.25) {};
		\node [style=none] (3) at (-0.75, -0.25) {};
		\node [style=none] (4) at (1.25, -0.25) {};
		\node [style=none] (5) at (0, 0.75) {};
		\node [style=none] (6) at (-0.75, -1) {};
		\node [style=none, yshift=-1.2pt] (7) at (0, 0.25) {$\cap$};
		\node [style=none] (8) at (0.75, -0.25) {};
		\node [style=none] (9) at (0.75, -1) {};
	\end{pgfonlayer}
	\begin{pgfonlayer}{edgelayer}
		\draw [style=swap] (6.center) to (3.center);
		\draw [style=swap] (9.center) to (8.center);
		\draw [style=swap] (2.center) to (5.center);
		\draw [style=swap] (5.center) to (4.center);
		\draw [style=swap] (4.center) to (2.center);
	\end{pgfonlayer}
\end{tikzpicture}    
\]
which are such that the following equation holds: 
\beq\label{yanking}
\begin{tikzpicture}
	\begin{pgfonlayer}{nodelayer}
		\node [style=none] (0) at (-1, 0.5) {};
		\node [style=none] (1) at (-3, -1.75) {};
		\node [style=none] (2) at (-2.25, 1.5) {};
		\node [style=none] (3) at (-1.5, -0.5) {};
		\node [style=none, yshift=-1.2pt] (4) at (-2.25, 1) {$\cap$};
		\node [style=none] (5) at (-3, 0.5) {};
		\node [style=none] (6) at (-3.5, 0.5) {};
		\node [style=right label, yshift=-0.6mm] (7) at (-1.25, 0) {$A^*$};
		\node [style=none] (8) at (-1.5, 0.5) {};
		\node [style=none] (9) at (0, 1.75) {};
		\node [style=none] (10) at (-2, -0.5) {};
		\node [style=none, yshift=1.2pt] (11) at (-0.75, -1) {$\cup$};
		\node [style=none] (12) at (0, -0.5) {};  
		\node [style=none] (13) at (-0.75, -1.5) {};
		\node [style=none] (14) at (0.5, -0.5) {};
		\node [style=none] (15) at (1.5, 0) {$=$};
		\node [style=right label, yshift=-0.6mm] (16) at (2.75, 0) {$A$};
		\node [style=none] (17) at (2.5, -1.75) {};
		\node [style=none] (18) at (2.5, 1.75) {};
		\node [style=right label, yshift=-0.6mm] (19) at (0.25, 1.5) {$A$};
		\node [style=right label, yshift=-0.6mm] (20) at (-2.75, -1.5) {$A$};
	\end{pgfonlayer}
	\begin{pgfonlayer}{edgelayer}
		\draw [style=swap] (1.center) to (5.center);
		\draw [style=swap] (3.center) to (8.center);
		\draw [style=swap] (6.center) to (2.center);
		\draw [style=swap] (2.center) to (0.center);
		\draw [style=swap] (0.center) to (6.center);
		\draw [style=swap] (9.center) to (12.center);
		\draw [style=swap] (10.center) to (13.center);
		\draw [style=swap] (13.center) to (14.center);
		\draw [style=swap] (14.center) to (10.center);
		\draw [style=swap] (18.center) to (17.center);
	\end{pgfonlayer}
\end{tikzpicture}
\eeq
\end{definition}

The reason for  `anti' in the name of this kind of togetherness is the fact that when a theory of togetherness is both cartesian and anti-cartesian, then it is nothing but a theory of absolute death, i.e.~it describes a world in which nothing ever  happens. Indeed, we have:
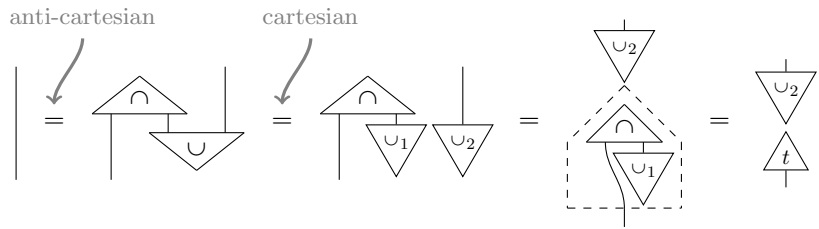
\[
\begin{tikzpicture}
	\begin{pgfonlayer}{nodelayer}
		\node [style=none] (0) at (-7.75, 0) {$=$};
		\node [style=none] (1) at (-8.75, -1.5) {};
		\node [style=none] (2) at (-8.75, 1.5) {};
		\node [style=point] (3) at (1.25, -0.5) {$\cup_1$};
		\node [style=none] (4) at (1.25, 0.25) {};
		\node [style=none] (5) at (3, 1.5) {};
		\node [style=point] (6) at (3, -0.5) {$\cup_2$};
		\node [style=none] (7) at (-5.25, -0.25) {};
		\node [style=none] (8) at (-2.75, -0.25) {};
		\node [style=none] (9) at (-3.25, 1.5) {};
		\node [style=none] (10) at (-4.75, 0.25) {};
		\node [style=none] (11) at (-3.25, -0.25) {};
		\node [style=none] (12) at (-4.25, 0.25) {};
		\node [style=none, yshift=-1.2pt] (13) at (-5.5, 0.75) {$\cap$};
		\node [style=none] (14) at (-6.75, 0.25) {};
		\node [style=none, yshift=1.2pt] (15) at (-4, -0.75) {$\cup$};
		\node [style=none] (16) at (-6.25, -1.5) {};
		\node [style=none] (17) at (-4.75, -0.25) {};
		\node [style=none] (18) at (-6.25, 0.25) {};
		\node [style=none] (19) at (-4, -1.25) {};
		\node [style=none] (20) at (-5.5, 1.25) {};
		\node [style=none] (21) at (-1.75, 0) {$=$};
		\node [style=none] (22) at (-0.75, 0.25) {};
		\node [style=none] (23) at (-0.25, -1.5) {};
		\node [style=none] (24) at (1.75, 0.25) {};
		\node [style=none, yshift=-1.2pt] (25) at (0.5, 0.75) {$\cap$};
		\node [style=none] (26) at (-0.25, 0.25) {};
		\node [style=none] (27) at (0.5, 1.25) {};
		\node [style=none] (28) at (4.75, 0) {$=$};
		\node [style=none] (29) at (6.25, -0.5) {};
		\node [style=none] (30) at (7.25, 0.5) {};
		\node [style=point] (31) at (7.25, 2) {$\cup_2$};
		\node [style=none] (32) at (7.75, -0.5) {};
		\node [style=point] (33) at (7.75, -1.25) {$\cup_1$};
		\node [style=none] (34) at (8.25, -0.5) {};
		\node [style=none] (35) at (6.75, -0.5) {};
		\node [style=none] (36) at (7.25, -2.25) {};
		\node [style=none] (37) at (7.25, 2.75) {};
		\node [style=none, yshift=-1.2pt] (38) at (7.25, 0) {$\cap$};
		\node [style=none] (39) at (7.25, 1) {};
		\node [style=none] (40) at (8.75, -2.25) {};
		\node [style=none] (41) at (5.75, -2.25) {};
		\node [style=none] (42) at (-1, 2.25) {};
		\node [style=label] (43) at (-1, 2.75) {\color{gray}cartesian};
		\node [style=none] (44) at (-1.75, 0.5) {};
		\node [style=label] (45) at (-7, 2.75) {\color{gray}anti-cartesian};
		\node [style=none] (46) at (-7.75, 0.5) {};
		\node [style=none] (47) at (-7, 2.25) {};
		\node [style=none] (48) at (9.75, 0) {$=$};
		\node [style=point, yshift=1mm] (49) at (11.5, 0.75) {$\cup_2$};
		\node [style=copoint, yshift=-1mm] (50) at (11.5, -0.75) {$t$};
		\node [style=none, yshift=1mm] (51) at (11.5, 1.5) {};
		\node [style=none, yshift=-1mm] (52) at (11.5, -1.5) {};
		\node [style=none] (53) at (7.25, -2.75) {};
		\node [style=none] (54) at (8.75, -0.5) {};
		\node [style=none] (55) at (5.75, -0.5) {};
	\end{pgfonlayer}
	\begin{pgfonlayer}{edgelayer}
		\draw [style=swap] (2.center) to (1.center);
		\draw [style=swap] (4.center) to (3);
		\draw [style=swap] (5.center) to (6);
		\draw [style=swap] (16.center) to (18.center);
		\draw [style=swap] (17.center) to (10.center);
		\draw [style=swap] (14.center) to (20.center);
		\draw [style=swap] (20.center) to (12.center);
		\draw [style=swap] (12.center) to (14.center);
		\draw [style=swap] (9.center) to (11.center);
		\draw [style=swap] (7.center) to (19.center);
		\draw [style=swap] (19.center) to (8.center);
		\draw [style=swap] (8.center) to (7.center);
		\draw [style=swap] (23.center) to (26.center);
		\draw [style=swap] (22.center) to (27.center);
		\draw [style=swap] (27.center) to (24.center);
		\draw [style=swap] (24.center) to (22.center);
		\draw [style=swap] (32.center) to (33);
		\draw [style=swap] (37.center) to (31);
		\draw [style=swap, in=-90, out=90, looseness=1.00] (36.center) to (35.center);
		\draw [style=swap] (29.center) to (30.center);
		\draw [style=swap] (30.center) to (34.center);
		\draw [style=swap] (34.center) to (29.center);
		\draw [style=dashed] (41.center) to (40.center);
		\draw [style=pointer edge, in=90, out=-90, looseness=0.75] (42.center) to (44.center);
		\draw [style=pointer edge, in=90, out=-90, looseness=0.75] (47.center) to (46.center);
		\draw [style=swap] (51.center) to (49);
		\draw [style=swap] (50) to (52.center);
		\draw [style=swap] (36.center) to (53.center);
		\draw [style=dashed] (39.center) to (54.center);
		\draw [style=dashed] (54.center) to (40.center);
		\draw [style=dashed] (39.center) to (55.center);
		\draw [style=dashed] (55.center) to (41.center);
	\end{pgfonlayer}
\end{tikzpicture}
\]
That is, the \em identity \em is a constant process, always outputting the state $\cup_2$, independent of what the input is.  And if that isn't weird enough, any arbitrary process  $f$ does the same:
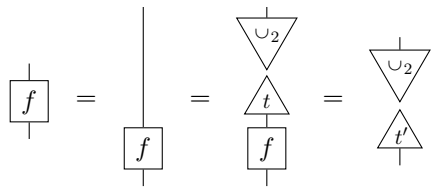
\[
\begin{tikzpicture}
	\begin{pgfonlayer}{nodelayer}
		\node [style=none] (0) at (2, 0) {$=$};
		\node [style=none] (1) at (-3, 2.5) {};
		\node [style=small box] (2) at (-3, -1.25) {$f$};
		\node [style=none] (3) at (-3, -2.25) {};
		\node [style=none] (4) at (-1.5, 0) {$=$};
		\node [style=small box] (5) at (0.25, -1.25) {$f$};
		\node [style=none] (6) at (0.25, -2.25) {};
		\node [style=none] (7) at (0.25, 2.5) {};
		\node [style=point] (8) at (0.25, 1.75) {$\cup_2$};
		\node [style=copoint] (9) at (0.25, 0) {$t$};
		\node [style=none, yshift=1mm] (10) at (3.75, 1.5) {};
		\node [style=none, yshift=-1mm] (11) at (3.75, -1.5) {};
		\node [style=point, yshift=1mm] (12) at (3.75, 0.75) {$\cup_2$};
		\node [style=copoint, yshift=-1mm] (13) at (3.75, -0.75) {$t'\!$};
		\node [style=small box] (14) at (-6, 0) {$f$};
		\node [style=none] (15) at (-4.5, 0) {$=$};
		\node [style=none] (16) at (-6, -1) {};
		\node [style=none] (17) at (-6, 1) {};
	\end{pgfonlayer}
	\begin{pgfonlayer}{edgelayer}
		\draw [style=swap] (1.center) to (2);
		\draw [style=swap] (2) to (3.center);
		\draw [style=swap] (5) to (6.center);
		\draw [style=swap] (7.center) to (8);
		\draw [style=swap] (10.center) to (12);
		\draw [style=swap] (9) to (5);
		\draw [style=swap] (13) to (11.center);
		\draw [style=swap] (17.center) to (14);
		\draw [style=swap] (14) to (16.center);
	\end{pgfonlayer}
\end{tikzpicture}
\]
Therefore,  any anti-cartesian theory of togetherness that involves some aspect of change cannot be cartesian, and hence will have interesting stuff emerging from togetherness.\footnote{Many more  properties of anti-cartesian togetherness can be found in \cite{CKbook}.}  
 
\section{Example 1: quantum theory}  
 
Anti-cartesian togetherness is a very particular alternative to cartesian togetherness (contra any theory that fails to be cartesian).  So one may wonder whether there are any interesting examples. And yes, there are!  One example is \em quantum entanglement \em in quantum theory.  That is in fact where the author's interest in anti-cartesian togetherness started \cite{LE1, AC1, Kindergarten}.\footnote{Independently, similar insights appeared in \cite{Baez, Kauffman}.} As shown in these papers, equation (\ref{yanking}) pretty much embodies the phenomenon of quantum teleportation \cite{Tele}.  The full-blown description of quantum teleportation goes as follows \cite{CPaqPav, CPer, CKbook}:
\ctikzfig{telefull}
It is not important to fully understand the details here. What is important is to note  that the bit of this diagram corresponding to equation (\ref{yanking}) is the bold wire which zig-zags through it:
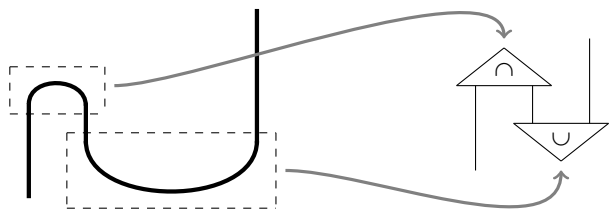
\ctikzfig{telefullbit}   
The thin wires and the boxes labelled $\widehat U$ are related to the fact that quantum theory is non-deterministic. By conditioning on particular measurement outcomes, teleportation simplifies to \cite{CKbook}:  
\ctikzfig{teledet}
Equality of the left-hand-side and of the right-hand-side follows directly from  equation (\ref{yanking}).    
While in this picture we eliminated quantum non-determinism by conditioning on a measurement outcome, there still is something very `quantum' going on here: Alice's (conditioned) measurement is nothing like a passive observation, but a highly non-trivial intervention that makes Alice's state $\rho$ appear at Bob's side:
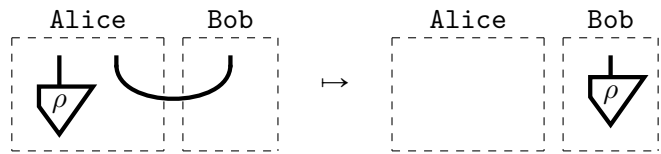
\ctikzfig{mapsto}  

Let's analyse more carefully what's going on here by explicitly distinguishing the top layer and the bottom layer of this diagram:  
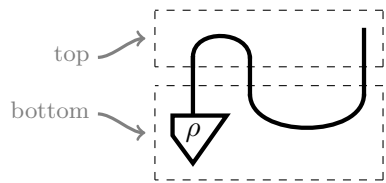
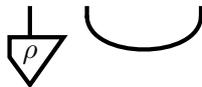
\ctikzfig{bottomtop}
The bottom part:  
\ctikzfig{bottom}    
consists of the state $\rho$ together with a  special \em $\cup$-state\em,           
while the top part:
\ctikzfig{top} 
includes the corresponding \em $\cap$-effect\em, as well as an output.  By making the bottom part and the top part interact, and, in particular, the $\cup$ and the $\cap$, the state $\rho$ ends up at the output of the top part. 

A more sophisticated variation on the same theme makes it much clearer which mechanism is going on here.       
Using equation (\ref{yanking}), the diagram:  
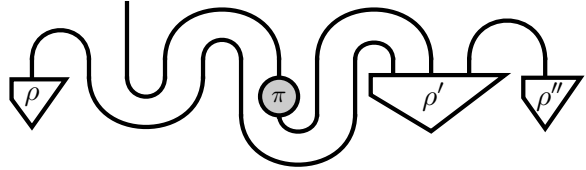
\ctikzfig{telecompl1pre}  
reduces to:
\[
\begin{tikzpicture}
	\begin{pgfonlayer}{nodelayer}
		\node [style=none] (0) at (-1, -0.75) {};
		\node [style=none] (1) at (1, -0.75) {};
		\node [style=none] (2) at (0, 1.5) {};
		\node [style=none] (3) at (0, 2.5) {};
		\node [style=gray phase ddot] (4) at (0, 1.5) {$\pi$};
		\node [style=dkpoint] (5) at (-1, -2) {$\rho$};
		\node [style=dkpoint] (6) at (1, -2) {$\rho''\!\!$};
		\node [style=large dmap] (7) at (0, -0.25) {$\rho'\!$};
	\end{pgfonlayer}
	\begin{pgfonlayer}{edgelayer}
		\draw [style=boldedge] (3.center) to (4);
		\draw [style=boldedge] (1.center) to (6);
		\draw [style=boldedge] (0.center) to (5);
		\draw [style=boldedge] (2.center) to (7);   
	\end{pgfonlayer}
\end{tikzpicture}\qquad\ \ \mbox{where}\qquad\ \ 
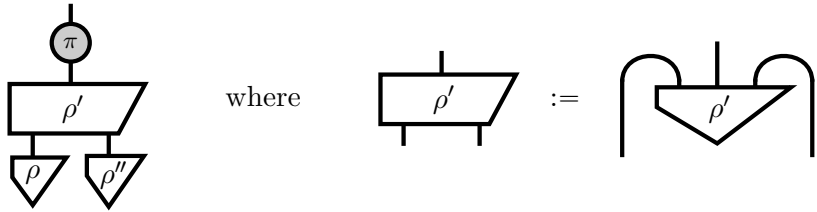
\begin{tikzpicture}
	\begin{pgfonlayer}{nodelayer}
		\node [style=none] (0) at (-4.25, -0.5) {};
		\node [style=none] (1) at (-2.25, -0.5) {};
		\node [style=none] (2) at (-3.25, 1.25) {};
		\node [style=none] (3) at (-3.25, 1.25) {};
		\node [style=none] (4) at (-4.25, -1.25) {};
		\node [style=none] (5) at (-2.25, -1.25) {};
		\node [style=large dmap] (6) at (-3.25, 0) {$\rho'\!$};
		\node [style=none] (7) at (0, 0) {$:=$};
		\node [style=none] (8) at (3, 0.25) {};
		\node [style=none] (9) at (1.5, 0.5) {};
		\node [style=none] (10) at (1.5, -1.5) {};
		\node [style=none] (11) at (4, 1.5) {};
		\node [style=none] (12) at (3, 0.5) {};
		\node [style=dkpoint, minimum width=1.3225 cm, inner sep=2 pt] (13) at (4, -0.25) {$\rho'\!$};
		\node [style=none] (14) at (6.5, -1.5) {};
		\node [style=none] (15) at (5, 0.25) {};
		\node [style=none] (16) at (5, 0.5) {};
		\node [style=none] (17) at (6.5, 0.5) {};
	\end{pgfonlayer}
	\begin{pgfonlayer}{edgelayer}
		\draw [style=boldedge] (1.center) to (5);
		\draw [style=boldedge] (0.center) to (4);
		\draw [style=boldedge] (2.center) to (6);
		\draw [style=boldedge] (12.center) to (8.center);
		\draw [style=boldedge] (9.center) to (10.center);
		\draw [style=boldedge, bend left=90, looseness=1.50] (9.center) to (12.center);
		\draw [style=boldedge] (11.center) to (13);
		\draw [style=boldedge] (16.center) to (15.center);
		\draw [style=boldedge] (17.center) to (14.center);
		\draw [style=boldedge, bend right=90, looseness=1.50] (17.center) to (16.center);
	\end{pgfonlayer}
\end{tikzpicture}  
\]
The grey dot labeled $\pi$ is some (at this point not important) unitary quantum operation \cite{CKbook}. Let us again consider the bottom and top parts:
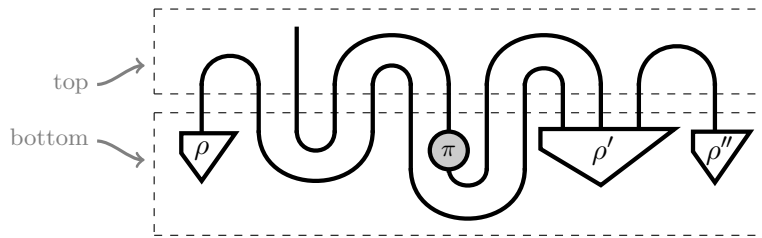
\ctikzfig{telecompl1}  
The top part is a far more sophisticated measurement consisting mainly of  $\cap$-s.  Also the bottom part is a lot more sophisticated, involving many $\cup$-s. These now cause a highly non-trivial interaction of the three states $\rho$, $\rho'$ and $\rho''$. Why we have chosen this particular example will become clear in the next section.  What is important to note is that the overall state and overall effect have to be chosen in a very particular way to create the desired interaction, similarly to an old-fashion telephone switchboard that has to be connected in a very precise  manner in order  to realise the right connection.

\section{Example 2: natural language meaning}    
 
Another example of anti-cartesian togetherness is the manner in which word meanings  interact in natural language!  Given that logic originated  in natural language, when Aristotle analysed arguments involving `and', `if...then', `or', etc., anti-cartesianness can be conceived as some new kind of logic!\footnote{A more detailed discussion is in \cite{QLog}.}  So what are $\cup$ and $\cap$  in this context?     

In order  to understand what $\cap$ is, we need to understand the mathematics of grammar.  The study of the mathematical structure of grammar has indicated that the fundamental things making up sentences are not  the words, but some atomic grammatical types, such as the noun-type and the sentence-type \cite{Ajdukiewicz, Bar-Hillel, Lambek0}.   The transitive verb-type is not an atomic grammatical type, but a composite made up of two noun-types and one sentence-type.  Hence, particularly interesting here is that atomic doesn't really mean smallest...

On the other hand, just like in particle physics where we have particles and anti-particles, the atomic types include  types as well as anti-types.  But unlike in particle physics, there are two kinds of anti-types, namely left ones and right ones.  This makes language even more non-commutative than quantum theory!

All of this becomes much clearer when considering an example.  Let $n$ denote the atomic \em noun\em-type and let  ${}^{-1}n$ and  $n^{-1}$ be the corresponding anti-types.  Let $s$ denote the atomic \em sentence\em-type.  Then the non-atomic \em transitive verb\em-type is ${}^{-1}n \cdot s \cdot n^{-1}$.  Intuitively, it is easy to understand why.  Consider a transitive verb, like `hate'.  Then, simply saying `hate' doesn't convey any useful information, until, we also specify `whom' hates `whom'.  That's exactly the role of the anti-types: they specify that in order to form a meaningful sentence, a noun is needed on the left, and a noun is needed on the right:
\[
\underbrace{Alice}_{n} \underbrace{hates}_{{}^{-1}n \cdot s \cdot n^{-1}}  \underbrace{Bob}_n
\]  
Then, $n$ and ${}^{-1}n$ cancel out, and so do $n^{-1}$ and $n$. What remains is $s$, confirming that `Alice hates Bob' is a grammatically well-typed sentence.  We can now depict the cancelations as follows:    
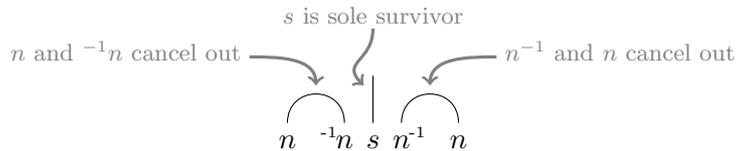
\ctikzfig{hates}
and bingo, we found $\cap$!    

While the mathematics of sentence structure has been explored now for some 80 years, the fact that $\cap$-s can account for grammatical structure is merely a 15 years old idea \cite{Lambek1}.  So what are the $\cup$-s?  That is an even more recent story  in which we were involved, and in fact, for which we took inspiration from the story of the previous section \cite{teleling}.    While $\cap$-s are about grammar,  $\cup$-s are about meaning.  

The  \em distributional \em paradigm for natural language meaning states that meaning can be represented by vectors in a vector space \cite{Schuetze}.  Until recently, grammatical structure was essentially ignored in doing so, and in particular, there was no theory for how to compute the meaning of a sentence, given the meanings of its words. Our new \em compositional distributional model of meaning \em of \cite{CSC} does exactly that.\footnote{...and has meanwhile outperformed other attempts in several benchmark natural language processing (NLP) tasks \cite{GrefSadr, KartSadr}.}  

In order to explain how this compositional distributional model of meaning works, let's get back to our example.  Since we have grammatical types around, the meaning vectors should respect grammatical structure, that is, the vectors representing compound types should themselves live in compound vector spaces.  So the string of vectors representing the word meanings of our example would look as follows:
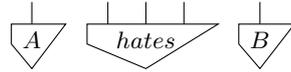
\ctikzfig{hates2}   
Now we want to put forward a new hypothesis:  
\begin{center}
\em Grammar is all about how word meanings interact.\em                   
\end{center}
Inspired by the previous section, this can be realised as follows:  
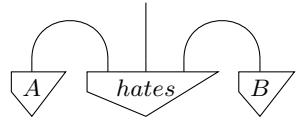
\ctikzfig{hates3}   
where the $\cap$-s are now interpreted in exactly the same manner as in the previous section.  And here is a more sophisticated example:
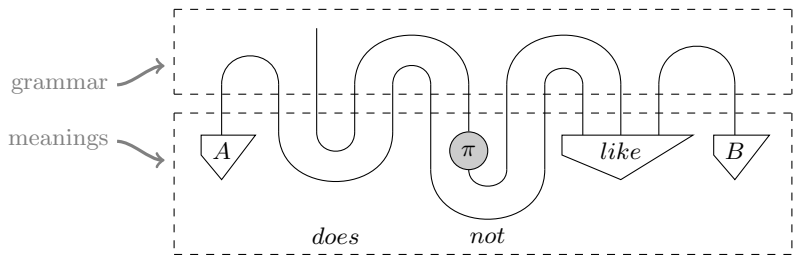
\ctikzfig{hatescompl}
where the $\pi$-labeled grey circle should now be conceived as negating meaning \cite{CSC}.  The grammatical structure  is here: 
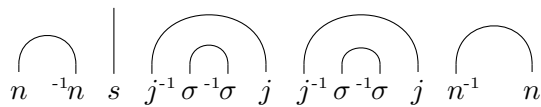
\ctikzfig{hatescomplgram}
It is simply taken from a textbook such as \cite{LambekBook}, the meanings of $Alice$, $likes$ and $Bob$ can be automatically generated from some corpus, while the meanings of $does$ and $not$ are just cleverly chosen to be \cite{PrelSadr, CSC}:  
\ctikzfig{hatescomplclever}
In the previous section we already saw that in this way we obtain:      
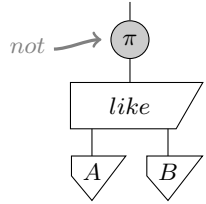
\ctikzfig{hatescompl2}
This  indeed captures the intended meaning:  
\[
not\left( like\left( Alice, Bob \right) \right)  
\]
where we can think of $like$ as being a predicate and $not$ as being itself.   

So an interesting new aspect of the last example is that some of the meaning vectors of words are simply cleverly chosen, and in particular, involve $\cup$-s.  Hence,  we genuinely exploit full-blown anti-cartesianess.  What anti-cartesianess does here is making sure that the transitive verb $likes$ `receives' $Alice$ as its object.  Note also how $not$ does pretty much the same as $does$, guiding word meanings through the sentence, with, of course, one very important additional task: negating the sentence meaning.  

 The cautious reader must of course have noticed that  in the previous section we used thick wires, while here we used thin ones. Also, the dots in the full-blown description of quantum teleportation, which represent classical data operations, have vanished in this section.  Meanwhile, thick wires as well as the dots all of these have acquired a vary natural role in a more refined model of natural language meaning. The dots allow to cleverly choose the meanings of relative pronouns \cite{FrobMeanI, FrobMeanII}:
\ctikzfig{relpron}
Thick wires (representing density matrices, rather than vectors \cite{CKbook})  allow to encode word ambiguity as \em mixedness \em \cite{RobinMSc, DimitriDPhil}. For example, the different meanings of the word $queen$ (a rock band, a person, a bee, a chess piece, or a drag ---).  Mixedness vanishes when providing a sufficient string of words that disambiguates that meaning, e.g.:       
\ctikzfig{queen}  
while in the case of:  
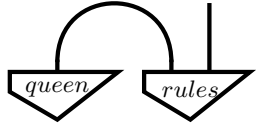
\ctikzfig{queen2}
we need more disambiguating words, since $queen$ can still refer to a person, a rock band, as well as a drag queen.  

\section{Meaning is everything}

The distributional model of meaning  \cite{Schuetze} is very useful in that it allows for automation, given a substantially large corpus of text.  However, from a conceptual point of view it is far from ideal.  So one may ask the question:
\begin{center}
\em What is meaning?\em 
\end{center}
One may try to play around with a variety of mathematical structures.  The method introduced in \cite{CSC} doesn't really depend on how one models meaning, as long as we stick to anti-cartesian togetherness, or something sufficiently closely related \cite{LambekvsLambek}.  It is an entertaining exercise to play around with the idea of what possibly could be the ultimate mathematical structure that captures meaning in natural language, until one realises that meaning in natural language truly encompasses \em everything\em.  Indeed, we use language to talk about everything, e.g.~logic, life, biology, physics, social behaviours, politics, so the ultimate model of meaning should encompass all of these fields.  So, a theory of meaning in natural language is actually a theory of everything! Can we make sense of the template introduced in the previous section for meaning in natural language, as one for ... everything?  

Let us first investigate, whether the general distributional paradigm can be specialised to the variety of subject domains mentioned above.  The manner in which the distributional model works, is that meanings are assigned relative to a fixed chosen set of context words.  The meaning vector of any word then arises by counting the number of occurrences of that word in the close neighbourhood of each of the context words, within a large corpus of text.  One can think of the context words as attributes, and the relative frequencies as the relevance of an attribute for the word.  Simply by specialising the context words and the corpus, one can specialise to a certain subject domain.  For example, if one is interested in social behaviours then the corpus could consist of social networking sites, and the context words could be chosen accordingly.   This pragmatic approach allows for quantitative analysis, just like the compositional distributional model of  \cite{CSC}.

Here's another example:
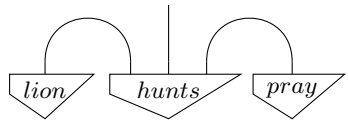
\ctikzfig{hunts}
Here the meaning  of pray could  include specification of the available pray, and then the meaning of the sentence would capture the survival success of the lion, given the nature of the available pray. All together, the resulting meaning is the result of the interaction between a particular hunter, a particular pray, and the intricacies of the hunting process, which may depend on the particular environment in which it is taking place.  It should be clear that again this situation is radically non-cartesian.

Of course, if we now consider the example of quantum theory from two sections ago, the analogues to grammatical types are system types i.e.~a specification of the kinds (incl.~quantity) of systems that are involved.   So it makes sense to refine the grammatical types according to the subject domain.  Just like nouns in physics would involve specification of the kinds of systems involved, in biology, for example, this could involve specification of species, population size, environment, availability of food etc.  Correspondingly, the top part would not just be restricted to grammatical interaction, but also domain specific interaction, just like in the case of quantum theory.    All together, what we obtain is the following picture:    
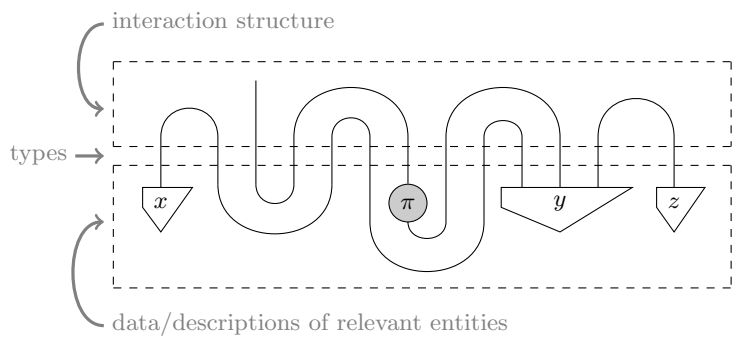
\ctikzfig{general}
as a (very rough) template for a theory of everything. 

%
%
%

\section*{Acknowledgements}

The extrapolation of meaning beyond natural language was prompted by having to give a course in a workshop on Logics for Social Behaviour, organised by Alexander Kurz and Alessandra Palmigiano at the Lorentz centre in Leiden.   The referee provided useful feedback---I learned a new word: `foo'.

\bibliographystyle{plain}
\bibliography{BC-final} 

\begin{thebibliography}{10}

\bibitem{AC1}
S.~Abramsky and B.~Coecke.
\newblock A categorical semantics of quantum protocols.
\newblock In {\em Proceedings of the 19th Annual IEEE Symposium on Logic in
  Computer Science (LICS)}, pages 415--425, 2004.
\newblock {a}rXiv:quant-ph/0402130.

\bibitem{Ajdukiewicz}
K.~Ajdukiewicz.
\newblock Die syntaktische konnexit\"at.
\newblock {\em Studia Philosophica}, 1:1--27, 1937.

\bibitem{Baez}
J.~C. Baez.
\newblock {Quantum quandaries: a category-theoretic perspective}.
\newblock In D.~Rickles, S.~French, and J.T. Saatsi, editors, {\em The
  Structural Foundations of Quantum Gravity}, pages 240--266. Oxford University
  Press, 2006.
\newblock {arXiv:quant-ph/0404040}.

\bibitem{Bar-Hillel}
Y.~Bar-Hillel.
\newblock A quasiarithmetical notation for syntactic description.
\newblock {\em Language}, 29:47--58, 1953.

\bibitem{benabou}
J.~Benabou.
\newblock Categories avec multiplication.
\newblock {\em Comptes Rendus des S\'eances de l'Acad\'emie des Sciences.
  Paris}, 256:1887--1890, 1963.

\bibitem{Tele}
C.~H. Bennett, G.~Brassard, C.~Crepeau, R.~Jozsa, A.~Peres, and W.~K. Wootters.
\newblock {Teleporting an unknown quantum state via dual classical and
  Einstein-Podolsky-Rosen channels}.
\newblock {\em Physical Review Letters}, 70(13):1895--1899, 1993.

\bibitem{brandao2011resource}
F.~G. S.~L. Brand{\~a}o, M.~Horodecki, J.~Oppenheim, J.~M. Renes, and R.~W
  Spekkens.
\newblock The resource theory of quantum states out of thermal equilibrium.
\newblock {\em Physical Review Letters}, 111:250404, 2013.

\bibitem{teleling}
S.~Clark, B.~Coecke, E.~Grefenstette, S.~Pulman, and M.~Sadrzadeh.
\newblock A quantum teleportation inspired algorithm produces sentence meaning
  from word meaning and grammatical structure.
\newblock arXiv:1305.0556, 2013.

\bibitem{LE1}
B.~Coecke.
\newblock The logic of entanglement. {A}n invitation.
\newblock Technical Report {RR-03-12}, {Department of Computer Science, Oxford
  University}, 2003.

\bibitem{Kindergarten}
B.~Coecke.
\newblock Kindergarten quantum mechanics.
\newblock In A.~Khrennikov, editor, {\em Quantum Theory: Reconsiderations of
  the Foundations III}, pages 81--98. AIP Press, 2005.
\newblock {a}rXiv:quant-ph/0510032.

\bibitem{QLog}
B.~Coecke.
\newblock The logic of quantum mechanics -- take {II}.
\newblock arXiv:1204.3458, 2012.

\bibitem{Gospel}
B.~Coecke.
\newblock An alternative {G}ospel of structure: order, composition, processes.
\newblock In C.~Heunen, M.~Sadrzadeh, and E.~Grefenstette, editors, {\em
  Quantum Physics and Linguistics. A Compositional, Diagrammatic Discourse},
  pages 1 -- 22. Oxford University Press, 2013.
\newblock arXiv:1307.4038.

\bibitem{CFS}
B.~Coecke, T.~Fritz, and R.~W. Spekkens.
\newblock A mathematical theory of resources.
\newblock {\em Information and Computation, to appear}, 2014.
\newblock arXiv:1409.5531.

\bibitem{LambekvsLambek}
B.~Coecke, E.~Grefenstette, and M.~Sadrzadeh.
\newblock Lambek vs. {L}ambek: Functorial vector space semantics and string
  diagrams for {L}ambek calculus.
\newblock {\em Annals of Pure and Applied Logic}, 164:1079--1100, 2013.

\bibitem{CKbook}
B.~Coecke and A.~Kissinger.
\newblock {\em Picturing Quantum Processes. A First Course in Quantum Theory
  and Diagrammatic Reasoning}.
\newblock Cambridge University Press, 2016.

\bibitem{CatsII}
B.~Coecke and {\'E}.~O. Paquette.
\newblock Categories for the practicing physicist.
\newblock In B.~Coecke, editor, {\em New Structures for Physics}, Lecture Notes
  in Physics, pages 167--271. Springer, 2011.
\newblock {a}rXiv:0905.3010.

\bibitem{CPaqPav}
B.~Coecke, {\'E}.~O. Paquette, and D.~Pavlovi{\'c}.
\newblock {Classical and quantum structuralism}.
\newblock In S.~Gay and I.~Mackie, editors, {\em Semantic Techniques in Quantum
  Computation}, pages 29--69. Cambridge University Press, 2010.
\newblock {a}rXiv:0904.1997.

\bibitem{CPer}
B.~Coecke and S.~Perdrix.
\newblock Environment and classical channels in categorical quantum mechanics.
\newblock In {\em Proceedings of the 19th EACSL Annual Conference on Computer
  Science Logic (CSL)}, volume 6247 of {\em Lecture Notes in Computer Science},
  pages 230--244, 2010.
\newblock Extended version: {a}rXiv:1004.1598.

\bibitem{CSC}
B.~Coecke, M.~Sadrzadeh, and S.~Clark.
\newblock Mathematical foundations for a compositional distributional model of
  meaning.
\newblock In J.~van Benthem, M.~Moortgat, and W.~Buszkowski, editors, {\em A
  Festschrift for Jim Lambek}, volume~36 of {\em Linguistic Analysis}, pages
  345--384. 2010.
\newblock ar{x}iv:1003.4394.

\bibitem{EPR}
A.~Einstein, B.~Podolsky, and N.~Rosen.
\newblock Can quantum-mechanical description of physical reality be considered
  complete?
\newblock {\em Physical review}, 47(10):777, 1935.

\bibitem{gour2008resource}
G.~Gour and R.~W. Spekkens.
\newblock The resource theory of quantum reference frames: manipulations and
  monotones.
\newblock {\em New Journal of Physics}, 10:033023, 2008.

\bibitem{GrefSadr}
E.~Grefenstette and M.~Sadrzadeh.
\newblock Experimental support for a categorical compositional distributional
  model of meaning.
\newblock In {\em The 2014 Conference on Empirical Methods on Natural Language
  Processing.}, pages 1394--1404, 2011.
\newblock ar{X}iv:1106.4058.

\bibitem{EntanglementResource}
R.~Horodecki, P.~Horodecki, M.~Horodecki, and K.~Horodecki.
\newblock Quantum entanglement.
\newblock {\em Reviews of Modern Physics}, 81:865--942, 2009.
\newblock arXiv:quant-ph/0702225.

\bibitem{DimitriDPhil}
D.~Kartsaklis.
\newblock {\em Compositional Distributional Semantics with Compact Closed
  Categories and Frobenius Algebras}.
\newblock PhD thesis, University of Oxford, 2014.

\bibitem{KartSadr}
D.~Kartsaklis and M.~Sadrzadeh.
\newblock Prior disambiguation of word tensors for constructing sentence
  vectors.
\newblock In {\em The 2013 Conference on Empirical Methods on Natural Language
  Processing.}, pages 1590--1601. ACL, 2013.

\bibitem{Kauffman}
L.~H. Kauffman.
\newblock Teleportation topology.
\newblock {\em Optics and Spectroscopy}, 99:227--232, 2005.

\bibitem{Lambek0}
J.~Lambek.
\newblock The mathematics of sentence structure.
\newblock {\em American Mathematics Monthly}, 65, 1958.

\bibitem{Lambek1}
J.~Lambek.
\newblock Type grammar revisited.
\newblock {\em Logical Aspects of Computational Linguistics}, 1582, 1999.

\bibitem{LambekBook}
J.~Lambek.
\newblock From word to sentence.
\newblock {\em Polimetrica, Milan}, 2008.

\bibitem{lieberman2013social}
M.~D. Lieberman.
\newblock {\em Social: Why our Brains are Wired to Connect}.
\newblock Oxford University Press, 2013.

\bibitem{RobinMSc}
R.~Piedeleu.
\newblock Ambiguity in categorical models of meaning.
\newblock Master's thesis, University of Oxford, 2014.

\bibitem{PrelSadr}
A.~Preller and M.~Sadrzadeh.
\newblock Bell states and negative sentences in the distributed model of
  meaning.
\newblock {\em Electronic Notes in Theoretical Computer Science},
  270(2):141--153, 2011.

\bibitem{Redei1}
M.~Redei.
\newblock Why {J}ohn von {N}eumann did not like the {H}ilbert space formalism
  of quantum mechanics (and what he liked instead).
\newblock {\em Studies in History and Philosophy of Modern Physics},
  27(4):493--510, 1996.

\bibitem{FrobMeanI}
M.~Sadrzadeh, S.~Clark, and B.~Coecke.
\newblock The {F}robenius anatomy of word meanings {I}: subject and object
  relative pronouns.
\newblock {\em Journal of Logic and Computation}, 23:1293--1317, 2013.
\newblock ar{X}iv:1404.5278.

\bibitem{FrobMeanII}
M.~Sadrzadeh, S.~Clark, and B.~Coecke.
\newblock The {F}robenius anatomy of word meanings {II}: possessive relative
  pronouns.
\newblock {\em Journal of Logic and Computation}, page exu027, 2014.

\bibitem{Schrodinger}
E.~Schr\"odinger.
\newblock Discussion of probability relations between separated systems.
\newblock {\em Cambridge Philosophical Society}, 31:555--563, 1935.

\bibitem{Schuetze}
H.~Sch{\"u}tze.
\newblock Automatic word sense discrimination.
\newblock {\em Computational linguistics}, 24(1):97--123, 1998.

\bibitem{Turing}
A.~M. Turing.
\newblock On computable numbers, with an application to the
  {E}ntscheidungsproblem.
\newblock {\em Proceedings of the London Mathematical Society}, 42:230--265,
  1937.

\bibitem{vN}
J.~von Neumann.
\newblock {\em Mathematische grundlagen der quantenmechanik}.
\newblock Springer-Verlag, 1932.
\newblock Translation, {\it Mathematical foundations of quantum mechanics},
  Princeton University Press, 1955.

\end{thebibliography}

\end{document}